\documentclass{article}

\usepackage{amsmath,amsfonts,bm}

\def\eqref#1{equation~\ref{#1}}
\def\1{\bm{1}}

\DeclareMathAlphabet{\mathsfit}{\encodingdefault}{\sfdefault}{m}{sl}
\SetMathAlphabet{\mathsfit}{bold}{\encodingdefault}{\sfdefault}{bx}{n}

\def\gG{{\mathcal{G}}}

\def\gX{{\mathcal{X}}}
\def\gY{{\mathcal{Y}}}

\usepackage[preprint]{tmlr}

\usepackage{url}            % simple URL typesetting
\usepackage{booktabs}       % professional-quality tables
\usepackage{amsfonts}       % blackboard math symbols
\usepackage{nicefrac}       % compact symbols for 1/2, etc.
\usepackage{microtype}      % microtypography
\usepackage{xcolor}         % colors

\usepackage{wrapfig}
\usepackage{subcaption}
\usepackage{comment}
\usepackage{multirow}
\usepackage{dirtytalk}
\usepackage{colortbl}
\usepackage{arydshln}
\usepackage{lipsum}
\usepackage{authblk}
\usepackage[ruled, linesnumbered]{algorithm2e}

\SetKwInput{KwInput}{Require} % Set the Input
\SetKwInput{KwOutput}{Return} % set the Output
\SetKwComment{Comment}{/* }{ */}
\definecolor{darkgreen}{RGB}{47, 135, 91} 
\definecolor{commentcolor}{RGB}{110,154,155}
\definecolor{LightCyan}{rgb}{0.88,1,1}
\definecolor{tabPurple}{rgb}{0.87, 0.87, 0.87}
\definecolor{tabPurple2}{rgb}{0.933, 0.937, 1}
\definecolor{tabOrange}{rgb}{0.99, 0.89, 0.79}
\definecolor{tabBlue}{rgb}{0.90, 0.96, 0.97}
\definecolor{tabBlue2}{rgb}{0.92, 0.95, 1}
\definecolor{tabGreen}{rgb}{0.86, 0.90, 0.85}
\definecolor{myyellow}{rgb}{0.98, 0.98, 0.82}
\definecolor{mygreen}{rgb}{0.94, 1.0, 0.94}
\definecolor{darkcyan}{RGB}{67, 117, 148}

\usepackage[colorlinks=true,citecolor=darkcyan,
            linkcolor=red]{hyperref}%
            
\usepackage{amsmath}
\usepackage{amssymb}
\usepackage{mathtools}
\usepackage{amsthm}

\theoremstyle{plain}

\theoremstyle{definition}

\theoremstyle{remark}

\newtheorem*{assumption*}{\assumptionnumber}
\providecommand{\assumptionnumber}{}
\makeatletter
\newenvironment{assumption}[2]
 {%
  \renewcommand{\assumptionnumber}{\textbf{\textup{Assumption #1}} \textup{(#2)}}%
  \begin{assumption*}%
  \protected@edef\@currentlabel{#1}%
 }
 {%
  \end{assumption*}
 }
\newcommand{\ourmethod}{\texttt{MAPL}}

\newcommand{\sayak}[1]{#1}

\title{\ourmethod{}: Model Agnostic Peer-to-peer Learning}

\author[1]{Sayak Mukherjee}
\author[2]{Andrea Simonetto}
\author[1,3]{Hadi Jamali-Rad}
\affil[1]{Computer Vision Lab, Delft University of Technology, The Netherlands}
\affil[2]{Unit\'e de Math\'ematiques Appliqu\'ees, ENSTA Paris, Institut Polytechnique de Paris, 91120 Palaiseau, France}
\affil[3]{Shell Global Solutions International B.V., Amsterdam, The Netherlands}

\begin{document}

\maketitle

\vspace{-0.2cm}
\begin{abstract}
\vspace{-0.2cm}
Effective collaboration among heterogeneous clients in a decentralized setting is a rather unexplored avenue in the literature. To structurally address this, we introduce \textbf{M}odel \textbf{A}gnostic \textbf{P}eer-to-peer \textbf{L}earning (coined as \ourmethod{}) a novel approach to simultaneously learn heterogeneous personalized models as well as a collaboration graph through peer-to-peer communication among neighboring clients. \ourmethod{} is comprised of two main modules: (i) local-level \textbf{P}ersonalized \textbf{M}odel \textbf{L}earning (\texttt{PML}), leveraging a combination of intra- and inter-client contrastive losses; (ii) network-wide decentralized \textbf{C}ollaborative \textbf{G}raph \textbf{L}earning (\texttt{CGL}) dynamically refining collaboration weights in a privacy-preserving manner based on local task similarities. Our extensive experimentation demonstrates the efficacy of \ourmethod{} and its competitive (or, in most cases, superior) performance compared to its centralized model-agnostic counterparts without relying on any central server. Our code is available and can be accessed here: \href{https://github.com/SayakMukherjee/MAPL}{https://github.com/SayakMukherjee/MAPL}.

\end{abstract}

\vspace{-0.3cm}
\section{Introduction}
\vspace{-0.3cm}

Achieving acceptable performance from deep learning models often hinges on the availability of a vast amount of data. In practice, a significant portion of this data is distributed across edge devices scattered globally. To avoid the cost of communication and honor the privacy of local devices, there has been a growing interest in research on \emph{decentralized learning} \citep{Beltran2022DecentralizedChallenges}. A possible approach to decentralized learning is \emph{Federated Learning} (FL) \citep{McMahan2017Communication-EfficientData}, wherein a central server acts as the coordinating agent among the clients. However, the central server introduces its own set of challenges, acting as a potential bottleneck for both communication and computation, with an inherent risk of disruption in the event of failure and/or malicious attacks \citep{Verbraeken2020ALearning, Wen2022AApplications}. From this perspective, and relying on a \emph{central server}, FL approaches are still centralized.

Seeking to address these limitations,  Peer-to-Peer (P2P) learning (also sometimes referred to as Decentralized Federated Learning)  \citep{Sun2022DecentralizedAveraging} has been proposed. Here, clients engage in direct communication with each other without reliance on a central server. This renders P2P learning more conducive to decentralized learning at scale. Further, a body of research has showcased that, under specific conditions, P2P learning converges to a \emph{consensus} model \citep{Hendrikx2020Dual-FreeReduction, Hendrikx2023AAlgorithms, Vogels2022BeyondLearning}, even surpassing its centralized (federated) counterparts \citep{Lian2017CanDescent}.

\textbf{Motivation.} Two inherent challenges in a decentralized learning setting are \emph{data heterogeneity} and \emph{model heterogeneity} across distributed clients. The same challenges apply to a centralized Federated Learning setting. To address data heterogeneity in a centralized setting, notable developments are Personalized Federated Learning (PFL) approaches \citep{Tan2022TowardLearning, Dong2023WeiAvg:Diversity, Tan2023PFedSim:Learning}, sometimes obtained through clustering \citep{Li2022TowardsMatching, Sattler2019ClusteredConstraints}. PFL diverges from the conventional approach of training a singular global model and instead focuses on the creation of personalized models tailored to individual or clustered client objectives. However, there has been limited exploration of PFL in P2P settings \citep{Zantedeschi2020FullyGraphs, Li2022LearningModels, Shi2023TowardsTraining}. 

To address model heterogeneity, \emph{model-agnostic} FL \citep{Regatti2022ConditionalLearning, Xu2023PersonalizedCollaboration} as well as PFL \citep{Jang2022FedClassAvg:Networks, Tan2021FedProto:Clients} approaches have been proposed allowing the network to operate in a more realistic setting with different client models. However, in the P2P setting, a widely used assumption is model homogeneity (same model, all clients) \citep{Sun2022DecentralizedAveraging, Shi2023ImprovingApproaches}, sometimes extended to scenarios where client models are distilled from the same larger backbone \citep{Ye2023HeterogeneousChallenges, Dai2022DisPFL:Training}. Although seemingly unrealistic, this assumption is deemed necessary to facilitate collaboration by aggregating client model parameters. On the other hand, in centralized FL, model heterogeneity has been addressed using knowledge distillation \citep{Zhang2021ParameterizedLearning}, HyperNetworks \citep{Shamsian2021PersonalizedHypernetworks} and by sharing class-wise feature aggregates with a central server \citep{Mu2021FedProc:Data, Tan2021FedProto:Clients}. While the former methods entail additional computation and communication overheads, the latter requires sharing sensitive information with the server, making it unfeasible to straightforwardly adapt to the P2P setting. Consequently, we pose the following pivotal question: \emph{can we train personalized models for heterogeneous clients in a P2P setting?}

\textbf{Contributions.} We introduce \textbf{M}odel \textbf{A}gnostic \textbf{P}eer-to-peer \textbf{L}earning, coined as \ourmethod{}. Operating within a model heterogeneous P2P setting (discussed in Section~\ref{sec:setting}), \ourmethod{} jointly learns personalized models and collaboration graph weights. At the local level, its \textbf{P}ersonalized \textbf{M}odel \textbf{L}earning (\texttt{PML}, in Section~\ref{sec:pfl}) module employs a combination of inter- and intra-client contrastive losses to learn separable class boundaries while mitigating the risk of learning biased representations due to an imbalanced label distribution. This is accomplished by leveraging shared information across neighboring clients via learnable prototype embeddings. Simultaneously, at the network level, the \textbf{C}ollaborative \textbf{G}raph \textbf{L}earning (\texttt{CGL}, in Section~\ref{sec:gl}) module dynamically refines the collaboration graph based on local task similarities. We apply a graph regularizer, offering fine-grained control over the sparsity of the learned collaboration network. Our main contributions can be summarized as follows:
\begin{itemize}
    \item We propose a novel approach: \ourmethod{}, to simultaneously learn \emph{heterogeneous personalized} models in a \emph{decentralized setting}.
    \item Under \ourmethod{}, we introduce two novel modules: \texttt{PML} incorporates learnable prototypes in a local contrastive learning setting to promote learning unbiased representations, and \texttt{CGL} dynamically discovers optimal collaboration weights in a privacy-aware decentralized manner.
    \item We conduct extensive experimentation with scenarios covering both data and model heterogeneity to substantiate the efficacy of \ourmethod{} (in Section~\ref{sec:eval}). We demonstrate that \ourmethod{} stays competitive with, or even outperforms, existing state-of-the-art model-agnostic centralized FL methods without relying on any central server. 
\end{itemize}

\vspace{-0.3cm}
\section{Related Work}
\vspace{-0.3cm}

\textbf{Personalized federated learning (PFL).} 
Recent studies have revealed that the centralized approach of FL suffers in data heterogeneous scenarios \citep{Hsieh2019TheLearning, Li2018FederatedNetworks, Li2021Model-ContrastiveLearning, Karimireddy2019SCAFFOLD:Learning}. To address this, instead of training a robust global model, personalized federated learning (PFL) aims at enhancing local performance through collaboration. The PFL approaches can be divided into five categories: parameter decoupling \citep{Arivazhagan2019FederatedLayers, Collins2021ExploitingLearning}, knowledge distillation \citep{Lin2020EnsembleLearning, He2020GroupEdge}, multitask learning \citep{Dong2023WeiAvg:Diversity, Tan2023PFedSim:Learning}, model interpolation \citep{Dinh2020PersonalizedEnvelopes, Li2021Ditto:Personalization} and clustering \citep{Li2022TowardsMatching, Jamali-Rad2022FederatedData}. For additional details, we refer the readers to \citet{Tan2022TowardLearning}. A prevalent assumption in the existing PFL literature is that all clients possess either identical network architectures or submodels pruned from a larger server model. This assumption significantly curtails the practical applicability of these approaches, given the diverse model architectures deployed across clients in real-world scenarios.

\textbf{Heterogeneous federated learning.} \sayak{These methods address system (a.k.a. model) heterogeneity next to data heterogeneity across clients \citep{Ye2023HeterogeneousChallenges, Liu2022CompletelyLearning, gao2022survey}}. pFedHN \citep{Shamsian2021PersonalizedHypernetworks} and FedRoD \citep{Chen2021OnClassification} utilize hypernetworks for personalization but introduce additional computational overhead. Knowledge distillation-based approaches, on the other hand, address system heterogeneity using a globally shared \citep{Zhang2021ParameterizedLearning} or generated datasets \citep{Zhu2021Data-FreeLearning}. Besides added communication overheads, distillation datasets might not match clients' local objectives. To overcome such shortcomings, \citet{Tan2021FedProto:Clients} propose to align the learned representations across clients using class-wise feature aggregates while training the final classification layers locally. Further, \citet{Regatti2022ConditionalLearning} proposes aggregation of the classifier layers in addition to class-wise feature centroids to improve the generalization performance of the clients. Conversely, \citet{Jang2022FedClassAvg:Networks} demonstrate that aggregating only the classifier layer suffices for representation alignment and improved generalization. Our proposed method resembles that of \citet{Jang2022FedClassAvg:Networks} and \citet{Tan2021FedProto:Clients} in using shared information among the clients while distinguishing itself by eliminating the need for a trusted central coordinating server.

\textbf{Decentralized learning.} Using a central server, traditional FL faces communication and computation bottlenecks, especially with a large number of clients, and is vulnerable to single-point failure attacks. To address this, in decentralized, federated learning (DFL) or peer-to-peer (P2P) learning, clients communicate only with immediate neighbors \sayak{determined by a communication graph}. Early work combined gossip averaging \citep{Xiao2007DistributedDeviation} and stochastic gradient descent utilizing doubly stochastic edge weights to obtain a global \say{consensus model}  \citep{Lian2017CanDescent, Jiang2017CollaborativeNetworks}.  Recent advancements include the application of decentralized federated averaging with momentum \citep{Sun2022DecentralizedAveraging} or with sharpness aware minimization (SAM) \citep{Shi2023ImprovingLearning}. However, these methods are not personalized for different clients. To address this gap, \citet{Vanhaesebrouck2016DecentralizedNetworks} propose to learn personalized models given a similarity graph as an input. Following this, \citet{Zantedeschi2020FullyGraphs} propose a joint objective for learning the collaboration graph in addition to the personalized models. Recent works propose a joint learning objective inferring task similarity through knowledge co-distillation \citep{Jeong2023PersonalizedDistillation} or meta-learning \citep{Li2022LearningModels}. In contrast, \citet{Dai2022DisPFL:Training} adopt a personalized sparse mask-based approach, and \citet{Shi2023TowardsTraining} consider a parameter-decoupled approach for personalization. \sayak{While our objective aligns with prior work in jointly learning personalized models and a collaboration graph, we focus on the more challenging \emph{decentralized model-heterogeneous} setting. Existing works in this direction rely on the knowledge distillation-based methods similar to centralized FL, which require sharing additional datasets \citep{li2021fedh2l} or models \citep{Kalra2023DecentralizedSharing, Khalil2024DFML:Learning} for alignment. However, as discussed earlier, such methods incur additional communication and computational overheads, which grow with the number of clients in the P2P setting. Instead, we propose to facilitate decentralized learning using shared feature prototypes, which drastically reduces the communication cost compared to sharing the model parameters. A concurrent work \citep{Li2024Prototype-BasedSystems} followed a similar approach to align heterogeneous clients in a time-varying topology. Our key difference stems from the capability of \ourmethod{} to identify relevant neighbors in a fully decentralized manner which is advantageous in two key aspects: (1) reduces the communication cost by sparsifying the collaboration graph \citep{Zantedeschi2020FullyGraphs, zhang2024energy} and (2) improves the overall performance by facilitating collaboration among clients with similar data distribution \citep{Sattler2019ClusteredConstraints, Li2022TowardsMatching}.} 

\vspace{-0.3cm}
\section{Problem Setting}  
\label{sec:setting}
\vspace{-0.3cm}

We consider a setting where $M$ clients solve a $K$-class classification task. Each client $i$ has access to a local data distribution $\mathcal{D}_i$ on $\mathcal{X} \times \mathcal{Y}$, where $\mathcal{X} = \mathbb{R}^d$ is the input space of dimension $d$ (with $d = \textup{width} \times \textup{height} \times \textup{nr. of channels}$ for image classification) and $\mathcal{Y} := \{1,..,K\}$ is the output space. Additionally, every client has a local model $h_i \in \mathbb{H}$ that maps samples from the input space $\mathcal{X}$ to the output space $\mathcal{Y}$. We stack all the client models in $\mathcal{H} = \{h_i\}_{i=1}^M$. Further, each client $i$ has access to $n_i$ samples $\{(x_q, y_q)\}_{q=1}^{n_i}$ drawn from $\mathcal{D}_i$, where $y_q \in \gY$ is the label associated with $x_q \in \gX$. Defining a local loss function $l: \mathcal{X} \times \mathcal{Y} \times \mathbb{H} \rightarrow \mathbb{R}$, the clients minimize the empirical risk:% as follows:
\begin{equation}
    \label{eq:opt2_2}
    \hat{\mathcal{L}}(\mathcal{H}) = \sum_{i\in[M]} \Big[\hat{l}(h_i) := \frac{1}{n_i}\sum_{q \in [n_i]} l(x_q, y_q; h_i)\Big].
\end{equation}
\textbf{Communication.} Each client can communicate with other clients in their local neighborhood $\mathcal{N}_i$ defined by a collaboration graph $\gG = \{[M], \bm{E}\}$ with $\bm{E}$ representing the edges along which messages can be exchanged. Each edge is associated with a weight $w_{ij}\in[0,1]$ capturing the similarity of local tasks between clients $i$ and $j$, with $w_{ij} = 0$ in the absence of an edge. We define the local neighborhood of a client $i$ as $\mathcal{N}_i = \{j\in [M]\setminus i | w_{ij} > 0\}$.
\begin{assumption}{1}{Mixing Matrix}
    \label{as:mm}
     We assume that the mixing matrix defined component-wise as $[\bm{W}]_{ij} = w_{ij}$, $\bm{W} \in [0,1]^{M \times M}$, is a row-stochastic matrix, i.e, $\bm{W1}_M = \bm{1}_M$.
\end{assumption}
\textbf{Client model.} Without loss of generality, the client model $h_i$ can be further decoupled into a feature extractor and classifier head. The feature extractor $f_{\theta_i}: \mathcal{X} \rightarrow \mathcal{Z}$ is a learnable network parameterized by $\theta_i$ which maps samples from the input space $\mathcal{X}$ to the latent space $\mathcal{Z} = \mathbb{R}^{d_z}$ and the classifier head $g_{\phi_i} :\mathcal{Z} \rightarrow \mathcal{Y}$ is a learnable network parameterized by $\phi_i$, which maps features from the latent space $\mathcal{Z}$ to the output space $\mathcal{Y}$. Therefore, a client model can be given by $h_i = f_{\theta_i} \circ g_{\phi_i}$, where $\circ$ is the function composition operator.
\begin{assumption}{2}{Model Heterogeneity}
     \label{as:mh}
     Given the decoupled client model and for the sake of a broader application, we assume the feature extractor $f$ can be different for every client, i.e., $\theta_i \neq \theta_j, \forall i,j \in [M],\, i \neq j$. However, the dimension of latent space $d_z$, and the classification head $g$ are considered to be the same across all clients \citep{Jang2022FedClassAvg:Networks, Tan2021FedProto:Clients}. \sayak{It can be easily achieved in practice by the addition of a fully connected or convolutional layer at the end of the heterogeneous backbone, which maps the output to the latent space with dimension $d_z$.}
\end{assumption}

\vspace{-0.3cm}
\section{Methodology}
\label{sec:mapl}
\vspace{-0.3cm}

We propose \textbf{M}odel \textbf{A}gnostic \textbf{P}eer-to-Peer \textbf{L}earning (\ourmethod), a novel method for learning \emph{heterogeneous personalized} models in a \emph{decentralized} setting while concurrently \emph{identifying neighbors} with similar tasks for effective collaboration. We first discuss the two subproblems: (i) peer-to-peer learning of personalized models (Section~\ref{sec:pfl}) and (ii) learning the collaboration graph (Section~\ref{sec:gl}). Next, in Section~\ref{sec:opt_prob} we elucidate how \ourmethod{} jointly addresses them in an \emph{alternating} fashion.

% \vspace{-0.3cm}
\subsection{Peer-to-Peer Learning of Personalized Models}
\label{sec:pfl} 
\vspace{-0.2cm}

\begin{figure*}[t]
    \centering
    \includegraphics[width=0.8\textwidth]{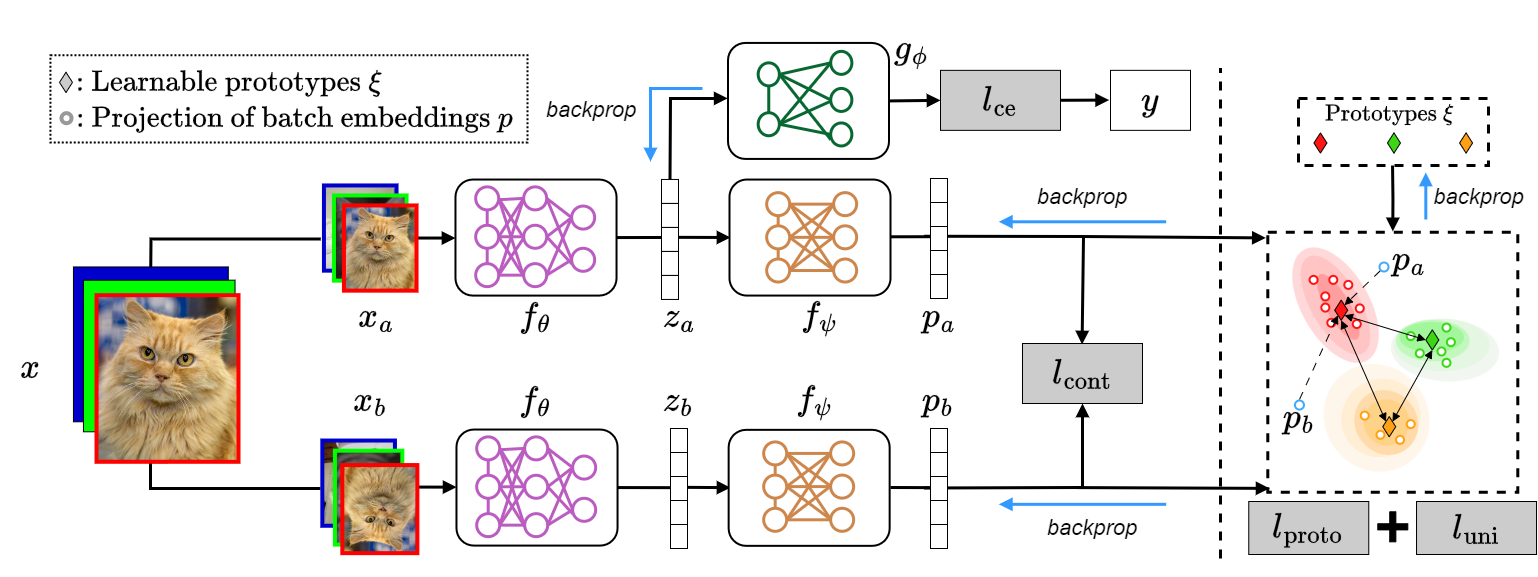}
    \caption{Illustration of the local training methodology of \ourmethod.}
    \label{fig:mapl_local}
    \vspace{-12pt}
\end{figure*}

 Given collaboration weights in $\bm{W}$, we focus on collaborative \textbf{P}ersonalized \textbf{M}odel \textbf{L}earning (\texttt{PML}) as summarized in Algorithm~\ref{algo:mapl}. Owing to model heterogeneity, the local models $h_i$ cannot be aggregated to facilitate collaboration. To form client relationships, we propose to modify the generic empirical risk in Eq.~\ref{eq:opt2_2} by adding a regularization term. This term depends on a piece of pertinent global information $\xi$ learned jointly by the clients and aggregated over the neighborhood $\mathcal{N}_i$ using the weights $\bm{W}$. As such, we formulate the regularized loss as:
\begin{equation}
    \label{eq:opt3_2}
    \Tilde{\mathcal{L}}(\mathcal{H}, \xi) = \sum_{i\in[M]} \Tilde{l}(h_i, \xi) := \hat{l}(h_i) + \mathcal{R}(h_i, \xi),
\end{equation}
where $\xi$ represents a set of learnable class-wise prototypes, which is further elaborated later in this section.
By minimizing $\Tilde{\mathcal{L}}(\mathcal{H}, \xi)$, we jointly learn the personalized models $h_i$ (using $\hat{l}(h_i)$) as well as improve upon the generalization capability of the local models by leveraging information $\xi$ gathered over the neighborhood $\mathcal{N}_i$ (using $\mathcal{R}(h_i, \xi)$). For $\hat{l}(h_i)$, we propose a combination of a contrastive loss $l_{\mathrm{cont}}$ and a cross-entropy loss $l_{\mathrm{ce}}$ to avoid biased local representations due to imbalanced label distributions at each client:
\begin{equation}
    \label{eq:ps_1}
    \hat{l}(h_i) = l_{\mathrm{cont}}(h_i) + l_{\mathrm{ce}}(h_i)\ .
\end{equation}
\sayak{Recent studies \citep{Khosla2020SupervisedLearning, graf2021dissecting} have demonstrated the effectiveness of using contrastive loss compared to cross-entropy loss for learning representations in the presence of class imbalance. Using $l_{\mathrm{cont}}$ representations in the latent space tend to collapse to the vertices of a regular $K$-simplex, thereby ensuring better separation among the class-wise features. Therefore, it has also been successful in overcoming the challenges posed by data heterogeneity in decentralized learning \citep{Wang2023DoesSupervision, Jang2022FedClassAvg:Networks}.} Let $\zeta_a, \zeta_b \sim \mathcal{A}$ be two randomly sampled augmentations from a set of data augmentations $\mathcal{A}$. At client $i$, a data sample $x \sim \mathcal{D}_i$ is transformed into two views $x_a = \zeta_a(x),\, \text{and}\, x_b =  \zeta_b(x)$. As illustrated in Fig.~\ref{fig:mapl_local}, we adopt a symmetric Siamese architecture similar to \citet{Chen2020ARepresentations}. Next, feature maps $z_a = f_{\theta_i}(x_a),\, \text{and}\, z_b = f_{\theta_i}(x_b)$ are obtained by passing both the views through the feature extraction backbone $f_{\theta_i}$. Akin to \citet{Khosla2020SupervisedLearning} and \citet{Jang2022FedClassAvg:Networks}, we apply sample-to-sample supervised contrastive loss ($l_{\mathrm{cont}}$) on the projected views. So, we consider a projection network $f_{\psi_i}: \mathcal{Z} \rightarrow \mathcal{P}$ parameterized by $\psi_i$ that maps samples from the latent space $\mathcal{Z}$ to the projected space $\mathcal{P} = \mathbb{R}^{d_p}$. Without loss of generality, we consider the same dimension for the latent and projection spaces, i.e., $d_p = d_z$. The extracted features are then passed through the projection network $f_{\psi_i}$ to obtain the projections $p_a = f_{\psi_i}(z_a),\, \text{and}\, p_b = f_{\psi_i}(z_b)$ (line $2-4$, Algorithm~\ref{algo:mapl}). We apply sample-to-sample supervised contrastive loss ($l_{\mathrm{cont}}$) on the projected views. Considering $\mathcal{Q} = \{1, \cdots, 2n_i\}$ as the indexes of the two augmented views, the supervised contrastive loss can be expressed as:
\begin{equation*}
        \label{eq:cont} 
        l_{\mathrm{cont}}(h_i) = - \sum_{q \in \mathcal{Q}} \frac{1}{|\mathcal{X}^+(y_q)|} \sum_{r \in \mathcal{X}^+(y_q)} \log \frac{\exp(\texttt{cos}(\hat{p}_q, \hat{p}_r)/\tau)}{\sum_{m \in \mathcal{Q} \setminus q} \exp(\texttt{cos}(\hat{p}_q, \hat{p}_m)/\tau)},
\end{equation*}
where the set of \emph{positive samples} for label $y_q$ (samples with the same label) is given by $\mathcal{X}^+(y_q) \equiv \{r \in \mathcal{Q}\setminus q \,|\, y_r = y_q\}$, $\hat{p}_q = p_q / \|p_q\|_2$ is the $\ell_2$ normalized projection of $x_q$, and $\tau$ is the temperature hyperparameter. The classifier head $g_{\phi_i}$ is trained by applying a cross-entropy loss ($l_{\mathrm{ce}}$) on the class logits obtained by passing the extracted feature maps through $g_{\phi_i}$ as $- \sum\nolimits_{q\in\mathcal{Q}} y_q \log(g_{\phi_i}(z_q))$ (line $6$, Algorithm~\ref{algo:mapl}). Here, both the projection head $f_{\psi_i}$ and predictor head $g_{\phi_i}$ are multilayer perceptrons (MLPs). 

\begin{wrapfigure}{r}{0.5\textwidth}
\vspace{-1.5em}
\begin{minipage}{0.5\textwidth}
\IncMargin{1.2em} 
\begin{algorithm}[H]
\small
\SetAlgoLined
\DontPrintSemicolon
\SetNoFillComment
\KwInput{$h_i = \{f_{\theta_i}, f_{\psi_i}, g_{\phi_i}\}, \xi_i, E, \eta$}
\For{$\textup{epoch} \in [E]~\textup{and}~(x, y) \in \mathcal{D}_i$}{
        $x_a, x_b \leftarrow \zeta_a(x), \zeta_b(x)$ for $\zeta_a, \zeta_b \in \mathcal{A}$\\
        $z_a, z_b \leftarrow f_{\theta_i}(x_a), f_{\theta_i}(x_b)$\\
        $p_a, p_b \leftarrow f_{\psi_i}(z_a), f_{\psi_i}(z_b)$\\
        {Compute local loss (Eq.~~\ref{eq:ps_1}):}\;
        {\quad $\hat{l}(h_i) = l_{\mathrm{cont}}(h_i) + l_{\mathrm{ce}}(h_i)$} \\
        {Compute regularization term:(Eq.~~\ref{eq:ps_2}):}\;
        {\quad $\mathcal{R}(h_i, \xi_i) = l_{\mathrm{proto}}(h_i, \xi_i) + l_{\mathrm{uni}}(\xi_i)$}\\
        {Compute loss (Eq.~~\ref{eq:opt3_2}):}\;
        {\quad $\Tilde{l}(h_i, \xi_i) = \hat{l}(h_i) + \mathcal{R}(h_i, \xi_i)$}\\
        {Update local model and prototypes:}\;
        {\quad $h_i \leftarrow h_i - \eta \nabla_{h_i} \Tilde{l}$\;
        \quad $\xi_i \leftarrow \xi_i - \eta \nabla_{\xi_i} \Tilde{l}$}
    % }
}
\KwOutput{$h_i, \xi$}
\caption{\texttt{PML}}
\label{algo:mapl}
\end{algorithm}
\DecMargin{1.2em}
\end{minipage}
\end{wrapfigure}

Now, a key question is: \emph{how do we benefit from relevant neighbors in a model heterogeneous setting?} Inspired by \citet{Mu2021FedProc:Data}, we propose to further align learned representations at neighborhood level by extending the contrastive setting using a shared set of \emph{learnable} class-wise prototypes $\xi = \{\xi^1, \cdots \xi^K\}$, with $\xi^k \in \mathcal{P}$, as shown in the Fig.~\ref{fig:mapl_local}. Remember that we referred to this set of prototypes as the global information in Eq.~\ref{eq:opt3_2} relating clients using the regularization term $\mathcal{R}$. Thus, we extend $\mathcal{R}$ as:
\begin{equation}
    \label{eq:ps_2}
    \mathcal{R}(h_i, \xi) = l_{\mathrm{proto}}(h_i, \xi) + l_{\mathrm{uni}}(\xi),
\end{equation}
where $l_{\mathrm{uni}}$ is a loss function that ensures maximum separability among the prototypes from different classes to prevent degenerate solutions. Specifically, the intra-client sample-to-prototype contrastive loss ($l_{\mathrm{proto}}$) applied on the $\ell_2$ normalized projections at every client is given by:
\begin{equation*}
        \label{eq:proto}
        l_{\mathrm{proto}}(h_i, \xi)\! =\! \frac{-1}{|\mathcal{Q}|} \sum_{q \in \mathcal{Q}} \log \frac{\exp(\texttt{cos}(\hat{p}_q, \xi^{y_q})/\tau)}{\displaystyle\sum_{k \in [K]} \!\!\!\exp(\texttt{cos}(\hat{p}_q, \xi_k)/\tau)},
\end{equation*}
where $\xi^{y_q}$ is the prototype for class $y_q$. \sayak{Note that it is an adapted version of the InfoNCE\citep{Oord2018RepresentationCoding} loss where the prototype $\xi^{y_q}$ serves as the positive pair for the representation $\hat{p}_q$ of the sample $(x_q, y_q)$ belonging to the class $y_q$.}

\sayak{For training a local set of prototypes $\xi_i$, we start from a random initialization at each client $i$ and aggregate them periodically as a weighted sum over the local neighborhood as $\xi_i \gets \{w_{ii}\, \xi_i^k + \sum\nolimits_{j \in \mathcal{N}_i} w_{ij}\, \xi_j^k, \forall\, k\, \in [K] \}$ (line 8, Algorithm~\ref{algo:mapl_overall}).} This forces class-wise representations to be closer to the shared class prototypes, thus effectively avoiding overfitting the local data distribution. \sayak{In our study, we do not consider concept shifts among the clients, thus making the aggregation of class-wise prototypes a valid choice for facilitating collaboration. 
 
Finally, to lower the similarity among inter-class prototypes in $\xi_i$ , we use a uniformity loss}:
\begin{equation}
    \label{eq:uni}
    l_{\mathrm{uni}}(\xi_i) = \frac{1}{K} \sum_{k \in [K]} \sum_{r \in [K]\setminus k} \frac{\langle \xi_i^k,\, \xi_i^r\rangle}{\| \xi_i^k \|_2 \cdot \| \xi_i^r \|_2}.
\end{equation}

\subsection{Learning the Collaboration Graph (\texttt{CGL})}
\label{sec:gl}
\vspace{-0.2cm}

%\ourmethod{}: A network perspective. 
\begin{figure*}[t]
    \centering
    \includegraphics[width=0.8\textwidth]{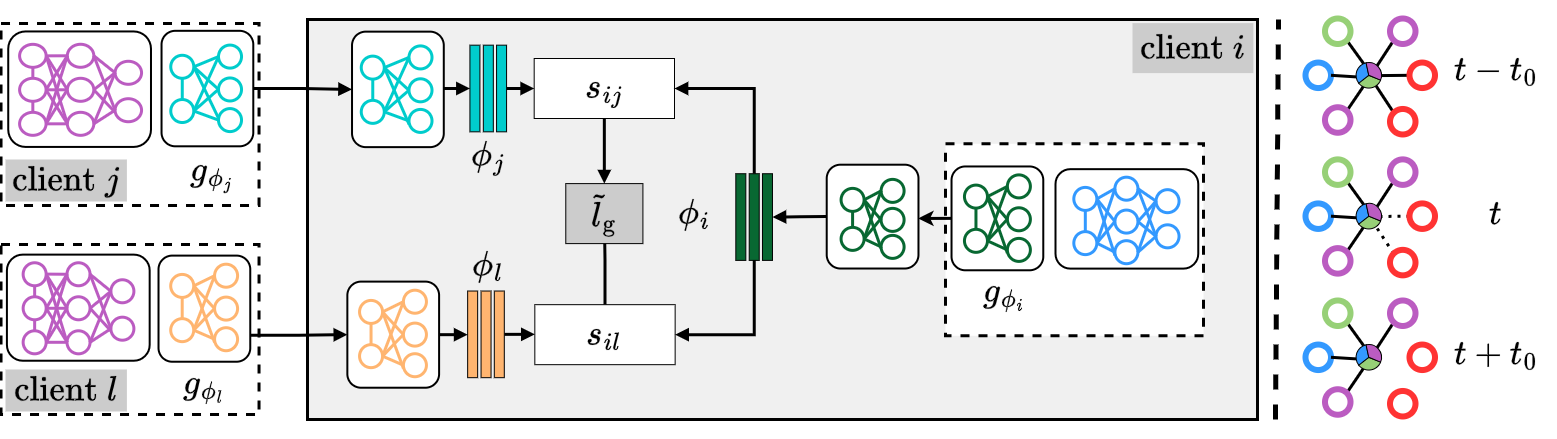}
    \caption{\textbf{(Left)} Computing similarity between clients. \textbf{(Right)} Evolution of collaboration graph: circles for clients, colors for classes.}
    \label{fig:mapl_gl}
\end{figure*}

We learn $\bm{W}$ based on an inferred similarity matrix $\bm{S} \in \mathbf{R}^{M \times M}$. To infer similarity without violating the data privacy, we propose to utilize the weight vector $\phi_i$ of the locally trained classifier heads $g_{\phi_i}$ as shown in Fig.~\ref{fig:mapl_gl} (line 3, Algorithm~\ref{algo:mapl_gl}). We argue that classifier heads, being representative of the true underlying data distribution, can be used to infer client similarity. 

\begin{figure}[t]
    \centering
    \includegraphics[width=0.75\linewidth]{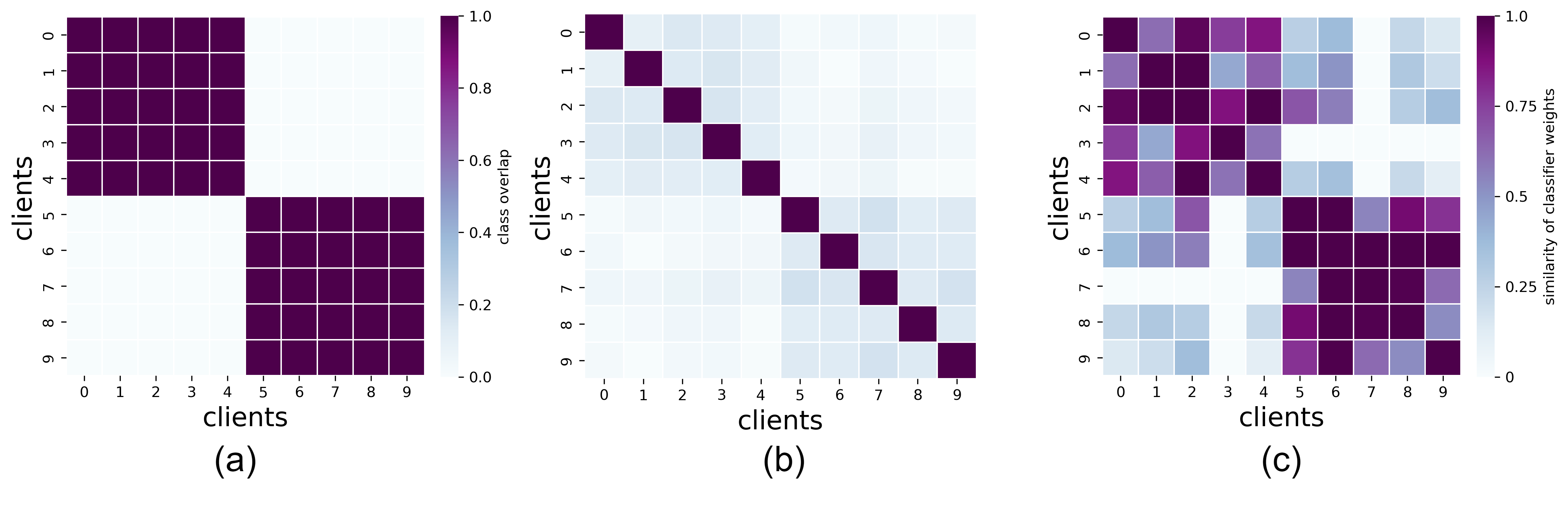}
    \vspace{-8pt}
    \caption{\textbf{(Left)} Data distribution and similarity of classifier weights trained without \textbf{(Middle)} and with \textbf{(Right)} $l_\text{proto}$ for aligning the class-wise feature representations.}
    \label{fig:class_sim}
    \vspace{-12pt}
\end{figure}

\sayak{A key concern arises in the applicability of using the classifier weights due to the presence of heterogeneous backbones. However, the \texttt{PML} algorithm employs shared prototypes $\xi_i$ to align class-wise feature embeddings across different model architectures. This ensures that the classifier learns to map features from corresponding positions in the latent space to output class labels, regardless of the underlying feature extractor backbone $f_\theta$. To validate our claim, we conducted an experiment with $10$ clients grouped into two clusters of $5$ each. Samples from classes $0-4$ of the Cifar10 dataset are assigned to the first cluster and classes $5-9$ to the second. Each client in the experiment is assigned a randomly chosen backbone from a set of four convolutional feature extraction networks. First, we train all clients without feature alignment in the latent space (without $l_\text{proto}$). The resulting similarity between classifier weights, as shown in Fig.~\ref{fig:class_sim}-(b), does not correlate with the underlying data distribution. However, upon aligning class-wise feature representations during training, the similarity of classifier weights serves as a proxy measure for client similarity, as illustrated in Fig.~\ref{fig:class_sim}-(c).}

Notably, the pFedSim \citep{Tan2023PFedSim:Learning} also adopts a similar approach in a centralized- setting \sayak{with homogeneous models}. Next to being decentralized, our work differs from pFedSim in that we learn the edge weights iteratively with the constraints of the mixing matrix (Assumption~\ref{as:mm}), as opposed to deterministically calculating it at every global round.

With classifier weight as $\phi_i = [\Phi_i^{(1)}, \cdots \Phi_i^{(K)}]$ with $\Phi_i^{(k)} \in \mathbb{R}^{d_z}, \forall k \in [K]$, inferred task similarity $\bm{s}_i$ is computed as the cosine similarities of the classifier weights:
\begin{equation*}
    \label{eq:sim}
    \setlength\abovedisplayskip{1.2em}
    \bm{s}_{i} = s(\phi_i, \mathcal{H}) = \{ - \langle \phi_i, \phi_j\rangle\; |\; \{f_{\theta_j}, g_{\phi_j}\} \in \mathcal{H},\; \forall j \in [M] \},
    \setlength\belowdisplayskip{1.2em}
\end{equation*}
where the $j$-th element $s_{ij}$ in the similarity vector $\bm{s}_i$ represents the similarity between client $i$ and $j$. 

\begin{wrapfigure}{r}{0.5\textwidth}
\begin{minipage}{0.5\textwidth}
\IncMargin{1.2em} 
\begin{algorithm}[H]
    \small
    \SetAlgoLined
    \DontPrintSemicolon
    \SetNoFillComment
    \Indm
    \KwInput{$\phi_i$, $\{\phi_j\}_{j=1}^{\mathcal{N}_i}$, $\mathcal{N}_i$, $\bm{w_i}$, $\mu_1$, $\mu_2$, $\eta$}
    \Indp
        \For{$\textup{step} \in [E_{\mathrm{g}}]$}{
            {Compute similarity with neighbors:}\; 
            {\quad $\bm{s}_i \gets s_{ij}\, \forall\, j \in \mathcal{N}_i$\;}
            {Compute graph learning loss (Eq.~\ref{eq:opt5_2_reg}):}\;
            {\quad $\Tilde{l}_{\mathrm{g}} \leftarrow\mu_1 \gamma\, (\bm{w}_i \odot \bm{s}_i) + \mu_2 \mathcal{R}_g(\bm{w}_i)$\;}
            {Update and project graph weights:}\;
            {\quad $\bm{w}_{i} \leftarrow { \bm w_{i}} - \eta \nabla_{{\bm w_{i}}} \Tilde{l}_{\mathrm{g}}$\;
            \quad ${\bm w}_{i} \leftarrow \texttt{Proj}(\bm{w}_{i})$\;}
        }
    \Indm
    \KwOutput{${\bm w_i}$}
    \Indp
    \caption{\texttt{CGL}}
    \label{algo:mapl_gl}
\end{algorithm}
\DecMargin{1.2em}
\end{minipage}
\vspace{-25pt}
\end{wrapfigure}

To assert a fully decentralized setting, each client $i$ only maintains a local weight vector $\bm{w}_i = [\bm{W}]_{i} \in [0,1]^{1 \times M}$, in contrast to the approaches discussed in \citep{Zantedeschi2020FullyGraphs, Jeong2023PersonalizedDistillation} which consider an overlay with a view of the entire collaboration network $\bm{W}$. The global graph learning loss can be given as:
\begin{equation}
    \setlength\abovedisplayskip{1.2em}
    \label{eq:opt4_2}
    \hat{\mathcal{L}}_{\mathrm{g}}(\bm{W}, \mathcal{H}) = \sum_{i\in[M]} \gamma\, (\bm{w}_i \odot \bm{s}_i),
    \setlength\belowdisplayskip{1.2em}
\end{equation}
where $\odot$ denotes element-wise multiplication and $\gamma \in [0,1]^{1\times M}$ is a proxy confidence vector with each element estimated as: $\gamma_i = |\mathcal{D}_i|/|\cup \mathcal{D}_j|, \forall j \in [M]$. \sayak{We propose to extend Eq.~\ref{eq:opt4_2} using a graph regularization term $\mathcal{R}_{\mathrm{g}}$ enforcing sparsity on the learned collaboration graph and reducing communications, thus, minimizing}:
\begin{equation}
    \label{eq:opt5_2_reg}
        \Tilde{\mathcal{L}}_{\mathrm{g}}(\bm{W}\!, \mathcal{H}) \!=\!\! \sum_{i\in[M]} \!\big[ \Tilde{l}_{\mathrm{g}}(\bm{w}_i,\! \bm{s}_i) \!:=\! \mu_1 \gamma\, (\bm{w}_i \odot \bm{s}_i) + \mu_2 \mathcal{R}_{\mathrm{g}}(\bm{w}_i)\!\big],
\end{equation}
where $\mu_1$ and $\mu_2$ are hyperparameters trading off a balance between the loss and the regularization. We adopt a regularization term $\mathcal{R}_{\mathrm{g}}$ inspired by \citep{Kalofolias2016HowSignals}:
\begin{equation}
    \mathcal{R}_{\mathrm{g}}(w_i) = \beta\|{\bm w_i}\|_2 - \log(\texttt{deg}(i) + \varepsilon),
\end{equation}
where $\texttt{deg}(i) = \sum_{j \in \mathcal{N}_i} w_{ij}$ is the degree of the node $i$ and $\varepsilon$ is a small positive constant. The log term prevents each client from being completely isolated from others while the $\ell_2$-norm, along with the hyperparameter $\beta$ controls the sparsity in the learned collaboration graph $\gG$. Finally, the \textbf{C}ollaborative \textbf{G}raph \textbf{L}earning (\texttt{CGL}), in Algorithm~\ref{algo:mapl_gl}, can be solved locally as a constrained optimization problem: 
\begin{equation}
    \label{eq:cgl_2}
    \begin{split}
        \operatorname*{minimize}_{\bm{w}_i}&\;\: \Tilde{l}_{\mathrm{g}}(\bm{w}_i, \bm{s}_i),\\
        \text{subject to}& \quad w_{ij} > 0\; \forall j \in [M],\; \bm{w}_i\bm{1}_M^T = 1.
    \end{split}
\end{equation}
Here, the non-negative constraint prevents negative values during the aggregation of prototypes, while the row-stochastic constraint ensures consensus. In practice, first, a gradient step is performed based on the graph loss $\Tilde{l}_{\mathrm{g}}$, followed by a projection onto a unit simplex (line 5, Algorithm~\ref{algo:mapl_gl}). We follow \citet[Algorithm 1]{Condat2016FastBall} for the projection function $\texttt{Proj}(.)$. Each client learns $\bm{w}_i$ over $E_{\mathrm{g}}$ steps in each global round.

\subsection{End to End Overview of \ourmethod}
\label{sec:opt_prob}
\vspace{-0.2cm}

\begin{wrapfigure}{r}{0.50\textwidth}
\vspace{-1.5em}
\begin{minipage}{0.50\textwidth}
\IncMargin{1.2em} 
\begin{algorithm}[H]
\small
\SetAlgoLined
\DontPrintSemicolon
\SetNoFillComment
\Indm
\KwInput{$M$, $\bm{w_i}$, $E$, $E_{\mathrm{g}}$, $T$, $T_{\textup{thr}}$, $\eta$, $\gamma$, $\{h_i\}_{i=1}^M = \{f_{\theta_i}, f_{\psi_i}, g_{\phi_i}\}_{i=1}^M, \{\xi_i\}_{i=1}^M$}
\Indp
\For{$t \in [T]~\textup{and}~i \in [M]$ in parallel}{
        %\Comment{update local models}
        \colorbox{myyellow}{Local step 1: $h_i, \xi_i \leftarrow \texttt{PML}(.)$}\\
        %\Comment{collaboration graph}
        \If{$t \geq \mathrm{T_{\textup{thr}}}$}{  
        \colorbox{mygreen}{Share classifier head with neighbors}\\
        \colorbox{myyellow}{Local step 2: $\bm{w}_{i} \gets \texttt{CGL}(.)$}}
        \colorbox{mygreen}{Share prototypes with neighbors}\\
        \colorbox{myyellow}{Local step 3: 
         $\xi_i \leftarrow  w_{ii}\, \xi_i + \sum_{j \in \mathcal{N}_i} w_{ij}\, \xi_j$}
}
\Indm
\KwOutput{$\{h_i\}_{i=1}^M$}
\Indp
\caption{\ourmethod{} \ \colorbox{myyellow}{Local step} \ \colorbox{mygreen}{Comm. step}}
\label{algo:mapl_overall}\vspace{0.45em}
\end{algorithm}
\DecMargin{1.2em}
\end{minipage}
\vspace{-20pt}
\end{wrapfigure}

An overview of \ourmethod{} is outlined in Algorithm~\ref{algo:mapl_overall}, which involves iterative joint learning of personalized models (referred to as \texttt{PML}, see Section~\ref{sec:pfl}) and collaboration weights (referred to as \texttt{CGL}, see Section~\ref{sec:gl}). In the \texttt{PML} step (line 2, Algorithm~\ref{algo:mapl_overall}, see also Fig.~\ref{fig:mapl_local}), each client jointly learns the model parameters $h_i$ and set of prototypes $\xi_i$ over $E$ epochs. Specifically, in this stage, we consider a combination of cross-entropy and contrastive loss to prevent biases in learned representations due to locally imbalanced label distributions, while inter-client alignment is achieved with a set of learnable prototypes regularized with a uniformity loss to prevent degenerate solutions. This is followed by \texttt{CGL} (line 3, Algorithm~\ref{algo:mapl_overall}, see also Fig.~\ref{fig:mapl_gl}), where the similarities between the clients are inferred based on the classifier layer parameters. The collaboration graph weight $\bm{W}$ is learnt using the locally inferred similarity $\bm{s}_i$ with a confidence vector $\gamma$. Note that \texttt{CGL} is performed after a fixed number of global epochs $T_{\textup{thr}}$ to ensure reliable training of classifier weight vectors. Lastly, the local set of prototypes $\xi_i$ is aggregated over the neighborhoods using $\bm{W}$ (line 8, Algorithm~\ref{algo:mapl_overall}).  We provide a Pytorch-like pseudo-codes in the Appendix.~\ref{sec:pymapl}.

\vspace{-0.3cm}
\section{Experimental Evaluation}
\label{sec:eval}
\vspace{-0.3cm}

In this section, we address the following two questions:

\textbf{[Q1]} How does \ourmethod{} perform against the \textit{state-of-the-art}?
\textbf{[Q2]} Can \ourmethod{} identify clients with similar classes?

\textbf{Benchmark dataset.} We study the effectiveness of \ourmethod{} using five widely used benchmark datasets in decentralized and federated learning: (i) CIFAR$10$, (ii) CINIC$10$, (iii) SVHN, (iv) MNIST and (v) FashionMNIST. Further details about these datasets can be found in the Appendix~\ref{sec:appendix-datasets}. 

\textbf{Data heterogeneity.} First, we consider label distribution skew, also often referred to as \emph{pathological heterogeneity} \citep{Jamali-Rad2022FederatedData}. More specifically, we group clients into $C$ clusters, with each cluster being assigned a subset of the available classes. We consider two scenarios: \textbf{Scenario $1$:} where clusters are allocated a disjoint subset of classes and, \textbf{Scenario $2$:} where an overlapping subset of classes are assigned to different clusters. Within each cluster, clients are further allocated $n_i = 300$ samples from each of the assigned classes. We also consider two additional settings where we simulate \textit{statistical heterogeneity} \citep{Kairouz2019AdvancesLearning} along with pathological heterogeneity. Here, clients within each cluster are randomly allocated between $n_i \in [100, 300]$ samples from each of the assigned classes, resulting in  \textbf{Scenario $3$} for clusters with disjoint classes, and \textbf{Scenario $4$} for clusters with overlapping classes.

\textbf{Model heterogeneity.} We consider two different settings: model \textbf{heterogeneous} (which is the core focus of this work) as well as model \textbf{homogeneous}, for the sake of baselining and completeness. In the heterogeneous setting, each client $i$ selects a feature extraction head $f_{\theta_i}$ randomly from a set of four models: GoogLeNet, ShuffleNet, ResNet$18$, and AlexNet, following \citet{Jang2022FedClassAvg:Networks}. Conversely, in the homogeneous setting, all clients use ResNet$18$ for feature extraction. Note that only \textit{model agnostic} approaches (e.g., \ourmethod) can be applied in both settings. 

\textbf{Training setup.} We use PyTorch for our implementations. Unless otherwise specified, our experiments consist of $T=400$ global communication rounds, with each client performing $E=1$ local epoch. Further implementation and hyperparameter details can be found in the Appendix~\ref{sec:appendix-impl}.

\subsection{Evaluation Results}
\label{sec:res}
\vspace{-0.2cm}

We report average test accuracies for $M =20$ clients along with their standard deviation, using $15$ test samples per class for each client. We compare the performance of \ourmethod{} against state-of-the-art model heterogeneous FL baselines \citep{Pillutla2022FederatedPersonalization, Jang2022FedClassAvg:Networks, Tan2021FedProto:Clients}. 

\textbf{[Q1]: Model heterogeneous setting.} We summarize the results for scenarios $1$ to $4$ in Tab.~\ref{tab:res_hetero}. Unlike the other methods, \ourmethod{} operates based on P2P communication without relying on a central server. Additionally, it utilizes graph learning to optimize edge weights as the training progresses. As can be seen, \ourmethod{} delivers competitive performance in all scenarios, even surpassing \emph{centralized} model-agnostic counterparts on challenging datasets such as CINIC$10$, SVHN, and CIFAR$10$ by $2-4\%$. We argue that this is due to an improved understanding of client relationships at the neighborhood level (using prototypes in \texttt{PML}) as well as at the global level (using \texttt{CGL}). \sayak{We compare \ourmethod{} against a set of model-agnostic FL baselines. Note that it is hard to compare the performance of \ourmethod{} with earlier works in model-agnostic P2P setting as they utilize additional information in the form of data \citep{li2021fedh2l} or models parameters \citep{Kalra2023DecentralizedSharing, Khalil2024DFML:Learning} for knowledge distillation. Therefore, we do not provide a comparison with these works.} For the sake of completeness, we also compare \ourmethod{} against a larger suite of prior art in a model-homogeneous setting in Appendix~\ref{sec:extended_res}, Tab.~\ref{tab:res_homo}. 

\begin{figure*}[t]
    \centering
    \includegraphics[width=\linewidth]{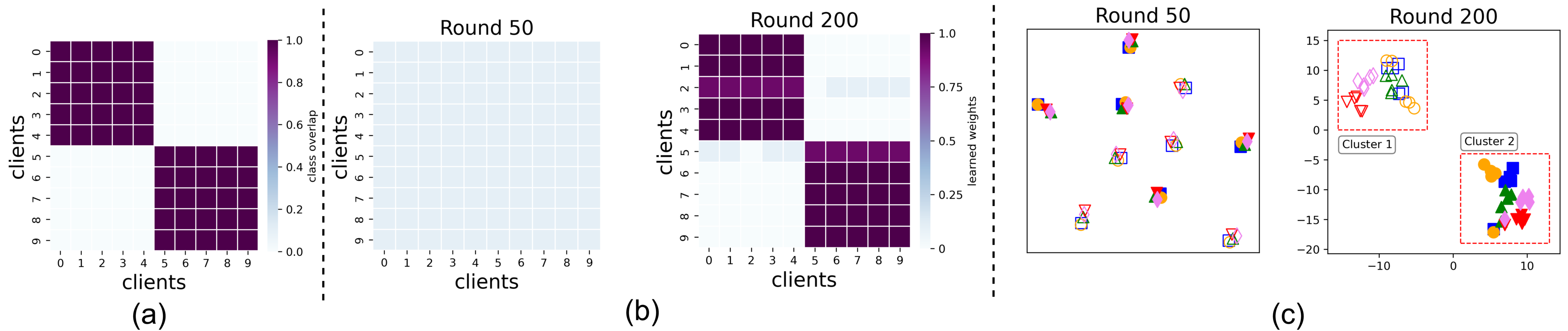}
    \caption{Client data distribution, \emph{learned} collaboration graph (part (b)) and latent embedding (part (c)) with $M=10$ for Scenario $1$.}
    \label{fig:col_graph}
    \vspace{-12pt}
\end{figure*}

\textbf{[Q2]: Learned collaboration graph.} We investigate the capacity of \ourmethod{} in discovering data distribution similarities among clients by visualizing the learned weight matrix $\bm{W}$ and the latent embedding $z$, with $M=10$ clients and $C=2$ clusters for Scenario $1$. \sayak{Clients $0$-$4$ are assigned to cluster $1$ with samples from classes $0$-$4$, while clients $5$-$9$ are assigned to cluster $2$ with samples from classes $5$-$9$. We show the percentage of overlapping classes between each pair of clients in Fig.~\ref{fig:col_graph}-(a)}. As can be seen in Fig.~\ref{fig:col_graph}-(b). \ourmethod{} successfully identifies the true clusters as communication rounds progress ($T$:$50$ $\rightarrow$ $200$), assigning equal weights to clients within the same cluster and almost zero weight to clients in the other cluster. Same can be seen in Fig.~\ref{fig:col_graph}-(c), where the latent embeddings at $T=50$ are spread across the $10$ clients, whereas at $T=200$ two distinct clusters with the right class assignments emerge, corroborating that \ourmethod{} successfully discovers the references data distribution in Fig.~\ref{fig:col_graph}-(a). Further results are provided in Appendix~\ref{sec:extended_res}.

\begin{table*}[t]
\caption{Average test accuracies in (\% ± std) for \emph{heterogeneous} models. Style: \textbf{best} and \underline{second best}.}
\label{tab:res_hetero}
  \centering
    \aboverulesep = 0pt
    \belowrulesep = 0pt
  \renewcommand{\arraystretch}{1.5}
  \resizebox{\linewidth}{!}{
  \begin{tabular}{l|c|c|c|c|c|c|c}
    \toprule
    \textbf{Method} & \textbf{Data Heterogeneity} & \textbf{Setting} & \cellcolor{tabPurple2}\textbf{CINIC$10$} & \cellcolor{tabPurple}\textbf{SVHN} & \cellcolor{tabOrange}\textbf{CIFAR$10$} & \cellcolor{tabBlue2}\textbf{MNIST} & \cellcolor{tabGreen}\textbf{FashionMNIST} \\
    \noalign{\vskip -1pt}
    \midrule
    \textbf{Local} & \texttt{Sc. 1} & \texttt{Local} & 64.13 ($\pm$08.46) & 70.60 ($\pm$11.63) & 70.60 ($\pm$09.26) & 97.40 ($\pm$02.75) & 91.27 ($\pm$04.96) \\
    \midrule
    \textbf{FedSim} \citep{Pillutla2022FederatedPersonalization} & \texttt{Sc. 1} & \texttt{Cent.} & 58.60 ($\pm$08.90) & 48.87 ($\pm$27.86) & 64.93 ($\pm$10.89) & 93.20 ($\pm$07.24) & 87.13 ($\pm$08.90) \\
    \textbf{FedClassAvg} \citep{Jang2022FedClassAvg:Networks} & \texttt{Sc. 1} & \texttt{Cent.} & \underline{70.53 ($\pm$09.43)} & \underline{73.40 ($\pm$11.10)} & \underline{74.67 ($\pm$09.65)} & 96.53 ($\pm$02.08) & \underline{91.00 ($\pm$04.23)} \\
    \textbf{FedProto} \citep{Tan2021FedProto:Clients} & \texttt{Sc. 1} & \texttt{Cent.} & 66.47 ($\pm$06.82) & 72.73 ($\pm$12.14) & 73.13 ($\pm$09.32) & \textbf{97.93 ($\pm$02.09)} & \textbf{92.40 ($\pm$04.74)} \\
    \noalign{\vskip -3pt}
    \hdashline
    \rowcolor{tabBlue}
    \textbf{\ourmethod{} (Ours)} & \texttt{Sc. 1} & \texttt{P2P} & \textbf{74.40 ($\pm$07.17)} & \textbf{76.07 ($\pm$10.36)} & \textbf{77.27 ($\pm$07.98)} & \underline{97.67 ($\pm$01.97)} & 90.80 ($\pm$05.13) \\
    \midrule \midrule
    \textbf{Local} & \texttt{Sc. 2} & \texttt{Local} & 49.81 ($\pm$08.59) & 68.43 ($\pm$12.96) & 58.67 ($\pm$08.71) & 95.71 ($\pm$02.86) & 86.52 ($\pm$03.77) \\
    \midrule
    \textbf{FedSim} \citep{Pillutla2022FederatedPersonalization} & \texttt{Sc. 2} & \texttt{Cent.} & 47.62 ($\pm$10.19) & 44.05 ($\pm$29.11) & 52.33 ($\pm$07.87) & 90.67 ($\pm$08.15) & 83.33 ($\pm$07.67) \\
    \textbf{FedClassAvg} \citep{Jang2022FedClassAvg:Networks} & \texttt{Sc. 2} & \texttt{Cent.} & \underline{57.14 ($\pm$08.08)} & \underline{70.14 ($\pm$12.98)} & \underline{62.86 ($\pm$08.43)} & 95.05 ($\pm$02.44) & 84.24 ($\pm$04.64) \\
    \textbf{FedProto} \citep{Tan2021FedProto:Clients} & \texttt{Sc. 2} & \texttt{Cent.} & 52.43 ($\pm$07.85) & 70.00 ($\pm$11.36) & 61.14 ($\pm$06.97) & \textbf{97.00 ($\pm$01.96)} & \underline{86.19 ($\pm$04.82)} \\
    \noalign{\vskip -3pt}
    \hdashline
    \rowcolor{tabBlue}
    \textbf{\ourmethod{} (Ours)} & \texttt{Sc. 2} & \texttt{P2P} & \textbf{58.76 ($\pm$06.40)} & \textbf{72.43 ($\pm$11.10)} & \textbf{65.43 ($\pm$08.40)} & \underline{96.62 ($\pm$02.03)} & \textbf{87.19 ($\pm$03.67)} \\
    \midrule \midrule
    \textbf{Local} & \texttt{Sc. 3} & \texttt{Local} & 58.67 ($\pm$06.38) & 62.93 ($\pm$14.70) & 65.40 ($\pm$08.94) & 96.00 ($\pm$04.04) & 89.67 ($\pm$05.43) \\
    \midrule
    \textbf{FedSim} \citep{Pillutla2022FederatedPersonalization} & \texttt{Sc. 3} & \texttt{Cent.} & 50.73 ($\pm$11.91) & 42.60 ($\pm$24.19) & 56.67 ($\pm$13.63) & 87.33 ($\pm$13.42) & 83.33 ($\pm$10.97) \\
    \textbf{FedClassAvg} \citep{Jang2022FedClassAvg:Networks} & \texttt{Sc. 3} & \texttt{Cent.} & \underline{65.40 ($\pm$08.15)} & \underline{66.53 ($\pm$15.21)} & \underline{69.53 ($\pm$09.98)} & 94.40 ($\pm$03.85) & 88.53 ($\pm$05.80) \\
    \textbf{FedProto} \citep{Tan2021FedProto:Clients} & \texttt{Sc. 3} & \texttt{Cent.} & 63.00 ($\pm$05.87) & 63.47 ($\pm$17.36) & 66.00 ($\pm$09.89) & \textbf{97.13 ($\pm$02.47)} & \textbf{91.07 ($\pm$04.77)} \\
    \noalign{\vskip -3pt}
    \hdashline
    \rowcolor{tabBlue}
    \textbf{\ourmethod{} (Ours)} & \texttt{Sc. 3} & \texttt{P2P} & \textbf{66.73 ($\pm$07.98)} & \textbf{71.07 ($\pm$13.12)} & \textbf{71.53 ($\pm$08.92)} & \underline{96.33 ($\pm$02.31)} & \underline{89.87 ($\pm$05.79)} \\
    \midrule \midrule
    \textbf{Local} & \texttt{Sc. 4} & \texttt{Local} & 46.43 ($\pm$07.52) & 60.29 ($\pm$14.75) & 53.52 ($\pm$08.03) & 94.09 ($\pm$04.35) & 83.62 ($\pm$04.24) \\
    \midrule
    \textbf{FedSim} \citep{Pillutla2022FederatedPersonalization} & \texttt{Sc. 4} & \texttt{Cent.} & 42.43 ($\pm$08.60) & 40.14 ($\pm$25.40) & 49.14 ($\pm$10.11) & 85.00 ($\pm$13.45) & 79.43 ($\pm$10.07) \\
    \textbf{FedClassAvg} \citep{Jang2022FedClassAvg:Networks} & \texttt{Sc. 4} & \texttt{Cent.} & \underline{52.29 ($\pm$05.61)} & \underline{63.67 ($\pm$13.33)} & \underline{57.29 ($\pm$07.34)} & 93.48 ($\pm$03.70) & 83.00 ($\pm$05.21) \\
    \textbf{FedProto} \citep{Tan2021FedProto:Clients} & \texttt{Sc. 4} & \texttt{Cent.} & 47.71 ($\pm$07.89) & 61.76 ($\pm$14.39) & 56.33 ($\pm$07.32) & \textbf{95.67 ($\pm$01.94)} & \textbf{85.14 ($\pm$03.56)} \\
    \noalign{\vskip -3pt}
    \hdashline
    \rowcolor{tabBlue}
    \textbf{\ourmethod{} (Ours)} & \texttt{Sc. 4} & \texttt{P2P} & \textbf{53.76 ($\pm$07.80)} & \textbf{67.24 ($\pm$13.98)} & \textbf{59.52 ($\pm$08.43)} & \underline{94.43 ($\pm$03.34)} & \underline{84.09 ($\pm$04.84)} \\
    \bottomrule
  \end{tabular}
  }
\end{table*}

% \vspace{-0.3cm}
\subsection{Ablation Studies}
\vspace{-0.3cm}

We investigate the performance of \ourmethod{} by varying numbers of clients ($M$) as well as analyze the impact of its key components in both \texttt{PML} and \texttt{CGL}. For all ablations, we consider the model-heterogeneous setting with $M=20$ clients and CIFAR$10$ dataset for Scenario $1$ and $2$.

\begin{wraptable}{r}{0.5\textwidth}
  \vspace{-12pt}
  \caption{Ablation of main components, (Acc.\% $\pm$ std).}
  \label{tab:ab1}
  \centering
  \aboverulesep = 0pt 
  \belowrulesep = 0pt
  \renewcommand{\arraystretch}{1.2}
  \resizebox{0.5\textwidth}{!}{%
  \scriptsize
  \begin{tabular}{@{}cc:ccc|c|c@{}}
    \toprule
    \textbf{$l_{\mathrm{ce}}$} & \textbf{$l_{\mathrm{proto}}$} & \textbf{$\Tilde{l}_{\mathrm{g}}$} & \textbf{$l_{\mathrm{cont}}$} & \textbf{$l_{\mathrm{uni}}$} & \textbf{Scenario 1} & \textbf{Scenario 2} \\
    \midrule
    \noalign{\vskip -1.2pt}
    \checkmark & \checkmark & - & - & - & 71.40 ($\pm$08.07) & 59.71 ($\pm$07.49)\\
    \checkmark & \checkmark & - & - & \checkmark & 71.40 ($\pm$08.32) & 59.38 ($\pm$09.08)\\
    \checkmark & \checkmark & - & \checkmark & - & 75.60 ($\pm$08.68) & 64.71 ($\pm$08.21)\\
    \checkmark & \checkmark & - & \checkmark & \checkmark & 75.67 ($\pm$08.86) & 65.19 ($\pm$07.26)\\
    \checkmark & \checkmark & \checkmark & - & - & 71.67 ($\pm$09.17) & 60.04 ($\pm$07.13)\\
    \checkmark & \checkmark & \checkmark & - & \checkmark & 72.40 ($\pm$08.25) & 60.05 ($\pm$08.01)\\
    \checkmark & \checkmark & \checkmark & \checkmark & - & 75.67 ($\pm$08.85) & 64.62 ($\pm$09.04)\\
    \checkmark & \checkmark & \checkmark & \checkmark & \checkmark & \textbf{77.27 ($\pm$07.98)} & \textbf{65.43 ($\pm$08.40)}\\
    \bottomrule
  \end{tabular}
  }
  \vspace{-10pt}
\end{wraptable}

\textbf{Ablation on main components.} We investigate the impact of the key loss terms within \ourmethod{}, namely $\Tilde{l}_g$, $l_{\mathrm{cont}}$, and $l_{\mathrm{proto}}$, by sequentially incorporating each term. Note that the cross-entropy loss ($l_{\mathrm{ce}}$) and sample-to-prototype loss ($l_{\mathrm{proto}}$) serve as foundational components, contributing to training the local classifier head and fostering collaboration among clients. As such, these two remain untouched as base performance. Initially, using the same collaboration weights for all clients,  our results in Tab.~\ref{tab:ab1} unveil the pivotal role played by the sample-to-sample contrastive loss ($l_{\mathrm{cont}}$), improving the performance by $~4\%$ and $~5\%$ in Scenario $1$ and $2$, respectively. Further, their combination with the uniformity loss ($l_{\mathrm{uni}}$) yields further additional accuracy boost, particularly pronounced in Scenario $2$. Incorporation of the graph learning loss ($\Tilde{l}_{\mathrm{g}}$) yields an additional $1-2\%$ enhancement in accuracy, culminating in the best performance in this setting. \sayak{Although the impact of $\Tilde{l}_{\mathrm{g}}$ can be considered a marginal improvement in the mean performance, an important factor to consider is the number of communications varies significantly among the two scenarios. Without the graph learning loss, we perform all-to-all communication, which scales with the number of nodes. Considering the number of nodes used in the experiments is 20, the total number of communications is $152,000$ over $400$ communication rounds. On the contrary, when using graph learning, after the initial warmup rounds $T_\textup{thr} = 100$ rounds, each client communicates with only the relevant neighbors. Thus, we consider the impact of $\Tilde{l}_{\mathrm{g}}$ to be significant as it allows us to obtain the same performance but with reduced communication overheads. We highlight this in Tab.~\ref{tab:comm_budget}, where we compare the need for the graph learning loss given while considering the number of communications. Lastly, we also report the results of our experiments with and without the $\Tilde{l}_{\mathrm{g}}$ using a fixed communication budget, where we can see in both scenarios 1 and 2 it results in improved performance.}

\begin{wraptable}{r}{0.5\textwidth}
\vspace{-12pt}
\caption{Ablation of graph learning (Acc.\% $\pm$ std).}
\label{tab:ab2}
  \centering
  \aboverulesep = 0pt 
  \belowrulesep = 0pt
  \renewcommand{\arraystretch}{1.2}
 \resizebox{0.5\textwidth}{!}{%
  \scriptsize
  \begin{tabular}{cccc|c|c}
    \toprule
    $\bm{S}$ & $\gamma$ & $\ell_1$ norm & $\ell_2$ norm & \textbf{Scenario 1} & \textbf{Scenario 2} \\
    \noalign{\vskip -1.2pt}
    \midrule
    \checkmark & - & - & \checkmark & 75.27 ($\pm$08.95) & 65.09 ($\pm$07.87)\\
     - & \checkmark & - & \checkmark & 76.20 ($\pm$07.73) & 64.38 ($\pm$08.26)\\
     \checkmark & \checkmark & \checkmark & - & 75.60 ($\pm$08.31) & 65.24 ($\pm$08.66)\\
     \checkmark & \checkmark & - & \checkmark & \textbf{77.27 ($\pm$07.98)} & \textbf{65.43 ($\pm$08.40)}\\
    \bottomrule
  \end{tabular}
  }
\end{wraptable}

\begin{table}[t]
      \caption{Performance vs Communication Budget (Acc.\% $\pm$ std).}
      \label{tab:comm_budget}
      \centering
      \aboverulesep = 0pt 
      \belowrulesep = 0pt
      \renewcommand{\arraystretch}{1.2}
      \resizebox{0.75\textwidth}{!}{%
      \scriptsize
      \begin{tabular}{l|c|c|c|>{\raggedleft\arraybackslash}p{1.8cm}l}
        \toprule
         & \textbf{Scenario 1} & \textbf{Scenario 2} & \textbf{Fixed Budget} & \multicolumn{2}{c}{\textbf{\# Comm.}}\\
         \noalign{\vskip -0.8pt}
        \midrule
        \noalign{\vskip -1.2pt}
        \textbf{\ourmethod{} w/o \texttt{CGL}} & 75.67 ($\pm$08.86) & 65.19 ($\pm$07.26) & - & 152,000 & \\
        \textbf{\ourmethod{}} & \textbf{77.27 ($\pm$07.98)} & \textbf{65.43 ($\pm$08.40)} & - & 101,000 & \multicolumn{1}{@{}l}{\tiny \textcolor{darkgreen}{($-34\%$)}}\\
        \midrule
        \rowcolor{tabBlue}
        \textbf{\ourmethod{} w/o \texttt{CGL}} &  72.53 ($\pm$06.97) & 61.48 ($\pm$09.04) & \checkmark & 65,000 & \\
        \rowcolor{tabBlue} \textbf{\ourmethod{}} & \textbf{75.20 ($\pm$07.32)} &  \textbf{63.09 ($\pm$08.07)} & \checkmark & 65,000 & \\
        \bottomrule
      \end{tabular}
      }
\end{table}

\textbf{Analysis of the graph learning objective.} Tab.~\ref{tab:ab2} illustrates the influence of the confidence vector ($\gamma$) and similarity matrix ($\bm{S}$) in the graph learning phase (\texttt{CGL}). It is interesting to note that in Scenario $2$, the similarity matrix has a greater impact than the confidence vector, whereas it is the opposite in Scenario $1$. This underscores the importance of inferring similarity between client data distributions in scenarios where partial class overlaps exist. However, irrespective of the scenario, the combination of the confidence vector and similarity matrix produces the optimal outcome. We further analyze the impact of using $\ell_1$ as opposed to the $\ell_2$ norm in the graph regularization term $\mathcal{R}_g$ for enforcing sparsity in the learned collaboration graph. From the results, it can be seen that the $\ell_2$ norm outperforms the $\ell_1$ norm in both scenarios, although the difference seems to be more pronounced for Scenario $1$ when compared to Scenario $2$. 

\begin{table}[tb]
\vspace{-0.5em}
\caption{Varying number of clients (Acc.\% $\pm$ std).}
\label{tab:var1}
  \centering
  \aboverulesep = 0pt 
  \belowrulesep = 0pt
  \renewcommand{\arraystretch}{1.2}
  \resizebox{0.8\linewidth}{!}{
  \begin{tabular}{@{}l|c|cc|cc@{}}
    \toprule
    \multirow{2}{*}{\textbf{Method}} & \multirow{2}{*}{\textbf{Setting}} & \multicolumn{2}{c|}{\textbf{Scenario 1}} & \multicolumn{2}{c}{\textbf{Scenario 2}} \\
    \cmidrule{3-6}
     &  & 10 clients & 20 clients & 10 clients & 20 clients \\
    \midrule
    \noalign{\vskip -1.4pt}
    \textbf{Local} & \texttt{Local} & 67.87 ($\pm$07.83) & 70.60 ($\pm$09.26) & 57.71 ($\pm$07.10) & 58.67 ($\pm$08.71) \\
    \midrule
    \noalign{\vskip -1.4pt}
    \textbf{FedSim} \hyperlink{cite.Pillutla2022FederatedPersonalization}{(2022)}  & \texttt{Cent.} & 63.20 ($\pm$08.83) &  64.93 ($\pm$10.89) & 54.19 ($\pm$07.65) & 52.33 ($\pm$08.71)\\
    \textbf{FedClassAvg} \hyperlink{cite.Jang2022FedClassAvg:Networks}{(2022)} & \texttt{Cent.} & 71.60 ($\pm$07.30) & 74.67 ($\pm$08.97) & 62.57 ($\pm$07.82) & 62.85 ($\pm$08.43)\\
    \textbf{FedProto} \hyperlink{cite.Tan2021FedProto:Clients}{(2021)} & \texttt{Cent.} & 69.60 ($\pm$09.25) & 73.13 ($\pm$09.32) & 61.62 ($\pm$07.67) & 61.14 ($\pm$06.97)\\
    \hdashline
    %\noalign{\vskip -1pt}
    \rowcolor{tabBlue}
    \textbf{\ourmethod{} (Ours)} & \texttt{P2P} & \textbf{73.20 ($\pm$07.31)} & \textbf{77.27 ($\pm$07.98)} & \textbf{63.62 ($\pm$08.40)} & \textbf{65.43 ($\pm$08.40)}\\
    \bottomrule
  \end{tabular}
  }
\end{table}

\textbf{Varying number of clients.} We consider $M=10$ and $M=20$ clients. As shown in Tab.~\ref{tab:var1}, \ourmethod{} outperforms all other model-agnostic baselines, regardless of their advantage of being centralized. When compared to $M=20$ clients, the margin for $M=10$ is about half (roughly $2\%$) in both scenarios, indicating a larger performance margin for \ourmethod{} as the network size grows. 

\textbf{Varying number of \texttt{CGL} steps.} Tab.~\ref{tab:cgle} demonstrated the performance of \ourmethod{} with varying numbers of \texttt{CGL} steps $E_{\mathrm{g}}$. It can be observed that for Scenario $1$, the performance declines initially before ramping up again, whereas for Scenario $2$, it merely keeps varying within a small range. Based on this empirical observation, we fixed the number of steps for \texttt{CGL} to be $E_{\mathrm{g}}=1$ for all the experiments. 

\begin{table}[t]
\caption{Varying number of \texttt{CGL} steps (Acc.\% $\pm$ std).}
\label{tab:cgle}
  \centering
  \aboverulesep = 0pt 
  \belowrulesep = 0pt
  \renewcommand{\arraystretch}{1.4}
  \resizebox{0.8\columnwidth}{!}{
  \begin{tabular}{@{}c|c|c|c|c|c@{}}
    \toprule
   \textbf{Scenario} & \textbf{1 epoch} & \textbf{2 epoch} & \textbf{5 epoch} & \textbf{10 epoch} & \textbf{20 epoch}\\
   \noalign{\vskip -1pt}
    \midrule
    1 & 77.27 ($\pm$07.98) & 74.40 ($\pm$09.72) & 76.13 ($\pm$07.47) & 76.00 ($\pm$08.53) & 76.73 ($\pm$07.98) \\
    2 & 65.43 ($\pm$08.40) & 64.52 ($\pm$08.67) & 65.28 ($\pm$08.68) & 64.52 ($\pm$08.96) & 65.52 ($\pm$08.62) \\   
    \bottomrule
  \end{tabular}
  }
\end{table}

% \vspace{-0.3cm}
\section{Concluding Remarks } 
\vspace{-0.3cm}

\begin{wraptable}{r}{0.5\textwidth}
\vspace{-12pt}
\caption{\small Communication Complexity Analysis}
\label{tab:comm}
  \centering
  \aboverulesep = 0pt 
  \belowrulesep = 0pt
  \renewcommand{\arraystretch}{1.32}
  \resizebox{0.5\textwidth}{!}{
  \begin{tabular}{@{}l|c|c@{}}
    \toprule
   \textbf{Method}  & \textbf{Setting} & \textbf{Complexity}\\
   \noalign{\vskip -1pt}
    \midrule
    \textbf{FedSim} \hyperlink{cite.Pillutla2022FederatedPersonalization}{(2022)} & \texttt{Cent.} & $2M\phi T$\\
    \textbf{FedClassAvg} \hyperlink{cite.Jang2022FedClassAvg:Networks}{(2022)} & \texttt{Cent.} & $2M\phi T$\\
    \textbf{FedProto} \hyperlink{cite.Tan2021FedProto:Clients}{(2022)} & \texttt{Cent.} & $2Md_p K T$\\
    \hdashline
    \textbf{FedAvg}  \hyperlink{cite.McMahan2017Communication-EfficientData}{(2022)} & \texttt{Cent.} & $2M(\theta + \phi)T$\\
    \textbf{FLT} \hyperlink{cite.Jamali-Rad2022FederatedData}{(2022)} & \texttt{Cent.} & $M\bm{W}_{\mathrm{enc}} + Md_pN_{\mathrm{emb}} + 2M(\theta + \phi)T$\\
    \textbf{pFedSim} \hyperlink{cite.Tan2023PFedSim:Learning}{(2022)} & \texttt{Cent.} & $2M(\theta + \phi)T$\\
    \hdashline
    \textbf{DFedAvgM} \hyperlink{cite.Sun2022DecentralizedAveraging}{(2022)} & \texttt{P2P} & $2N_{\mathrm{e}}(\theta + \phi)T$\\
    \hdashline
    \rowcolor{tabBlue}
    \textbf{\ourmethod{} (Ours)} & P2P & $2N_{\mathrm{e}}(d_p K + \phi)T$ \\
    \bottomrule
  \end{tabular}
  }
\end{wraptable}

\textbf{Computation complexity and convergence speed.} All aforementioned baselines have similar computational complexity at the client level, with the exception of FedClassAvg. \ourmethod{} and FedClassAvg incur slightly higher complexity, requiring twice the number of forward passes due to using $2$ views for each input sample. Further, \ourmethod{} utilizes additional computation at the client level for \texttt{CGL}. However, this overhead is negligible when compared to the cost of each local epoch, particularly when the number of epochs is large ($E \gg E_g$, with $E_g = 1$ in our experiments). On the other hand, being P2P, \ourmethod{} elevates the need for extra computation on the server side. We demonstrate in Appendix~\ref{sec:extended_res} that \ourmethod{} converges faster than all model-agnostic competitors.

\textbf{Communication complexity.} We compare the communication cost of \ourmethod{} with various baselines in Tab.~\ref{tab:comm}. Centralized model-agnostic approaches, namely FedSim \citep{Pillutla2022FederatedPersonalization}, FedClassAvg \citep{ Jang2022FedClassAvg:Networks}, and FedProto \citep{Tan2021FedProto:Clients}, prove to be the most efficient by sharing only the classifier head weights $\phi$ or the prototypes, instead of sharing parameters of the entire model ($\theta + \phi$). Further among the centralized baselines, FLT \citep{Jamali-Rad2022FederatedData} requires an additional one-off communication for sharing the encoder model ($\bm{W}_{\mathrm{enc}}$) with clients and retrieving $N_{\mathrm{emb}}$ local embeddings. In the P2P setting, the number of communications is inherently higher, with each client $i \in [M]$ communicating with all its neighbors $\mathcal{N}_i$. Nonetheless, \ourmethod{} incurs significantly lower communication cost compared to the P2P baseline DFedAvgM \citep{Sun2022DecentralizedAveraging} by not sharing backbone parameters $\theta$. Additionally, our approach incorporates sparsification using $\mathcal{R}_g$ in the graph learning phase (line 5 in Algorithm~\ref{algo:mapl_gl}), effectively minimizing costs.

\textbf{Broader impact and limitations.} \ourmethod{}, can operate in a fully \emph{decentralized model-heterogeneous} setting. This can potentially open the doors for further studies on decentralized FL in more practical settings. Further, \ourmethod{} can be utilized as an enhanced decentralized clustering approach (e.g., instead of decentralized $k$-means \citep{Fellus2013DecentralizedDatasets, Soliman2020DecentralizedCounters}). On the other hand, the current design of \ourmethod{} only allows for static (as opposed to dynamic) network connectivity graphs. \sayak{We will explore it in future work as it is beyond the scope of our current work. Additionally, our approach is based on the concept of sparsifying the collaboration graph, which entails starting from a fully connected mesh topology. Nonetheless, \ourmethod{} is an important stepping stone in the direction towards an exploration-based approach that gradually discovers relevant neighbors randomly querying a subset of connected neighbors in the network.} 

\section*{Acknowledgements}
This work is partially funded by the Shell.ai Innovation Program at Shell Global Solutions International B.V.

% \textbf{Do not} include acknowledgements in the initial version of
% the paper submitted for blind review.

% If a paper is accepted, the final camera-ready version can (and
% probably should) include acknowledgements. In this case, please
% place such acknowledgements in an unnumbered section at the
% end of the paper. Typically, this will include thanks to reviewers
% who gave useful comments, to colleagues who contributed to the ideas,
% and to funding agencies and corporate sponsors that provided financial

\clearpage

\textbf{Reproducibility Statement.} To help readers reproduce our experiments, we provide extensive descriptions of implementation details and algorithms. Architectural and training details are provided in Appendices~\ref{sec:appendix-arch} and~\ref{sec:appendix-train}, respectively, along with information on the applied data augmentations (Appendix~\ref{sec:appendix-augm}) and tested benchmark datasets (Appendix~\ref{sec:appendix-datasets}). The algorithms for \ourmethod{} along with the $\texttt{PML}$ and the $\texttt{CGL}$ modules are provided in both algorithmic (in Algorithms~\ref{algo:mapl}, \ref{algo:mapl_gl}, \ref{algo:mapl_overall}) and Pytorch-like pseudocode formats (in Algorithms~\ref{alg:mapl_pytorch}, \ref{alg:pml_pytorch}, \ref{alg:cgl_pytorch}). We have taken every measure to ensure fairness in our comparisons by following the most commonly adopted evaluation settings in the Federated Learning literature in terms of data heterogeneity. We also refer to the publicly available code base, if available, for comparing their performance with \ourmethod{}. Our codebase will also be made publicly available upon acceptance.

\textbf{Ethics Statement.} All benchmark datasets are publicly available and not directly subject to ethical concerns. We limited our study to $M=20$ clients due to the limitation of resources necessary for running multiple instances of large computer vision models such as ResNet$18$. Lastly, we are aware of the work Learning to Collaborate (L2C) \citep{Li2022LearningModels} for personalization in a P2P setting. We did not include it in this paper as there is no publicly available code, and we could not reliably reproduce the results as reported in the paper. Further, it does not support heterogeneous backbones, which is our main focus.


\bibliography{references, biblio}

\begin{thebibliography}{}

\bibitem[Arivazhagan et~al., 2019]{Arivazhagan2019FederatedLayers}
Arivazhagan, M.~G., Aggarwal, V., Singh, A.~K., and Choudhary, S. (2019).
\newblock {Federated Learning with Personalization Layers}.

\bibitem[Beltr{\'{a}}n et~al., 2022]{Beltran2022DecentralizedChallenges}
Beltr{\'{a}}n, E. T.~M., P{\'{e}}rez, M.~Q., S{\'{a}}nchez, P. M.~S., Bernal, S.~L., Bovet, G., P{\'{e}}rez, M.~G., P{\'{e}}rez, G.~M., and Celdr{\'{a}}n, A.~H. (2022).
\newblock {Decentralized Federated Learning: Fundamentals, State-of-the-art, Frameworks, Trends, and Challenges}.

\bibitem[Chen and Chao, 2021]{Chen2021OnClassification}
Chen, H.~Y. and Chao, W.~L. (2021).
\newblock {On Bridging Generic and Personalized Federated Learning for Image Classification}.
\newblock {\em ICLR 2022 - 10th International Conference on Learning Representations}.

\bibitem[Chen et~al., 2020]{Chen2020ARepresentations}
Chen, T., Kornblith, S., Norouzi, M., and Hinton, G. (2020).
\newblock {A Simple Framework for Contrastive Learning of Visual Representations}.
\newblock {\em 37th International Conference on Machine Learning, ICML 2020}, PartF168147-3:1575--1585.

\bibitem[Collins et~al., 2021]{Collins2021ExploitingLearning}
Collins, L., Hassani, H., Mokhtari, A., and Shakkottai, S. (2021).
\newblock {Exploiting Shared Representations for Personalized Federated Learning}.
\newblock {\em Proceedings of Machine Learning Research}, 139:2089--2099.

\bibitem[Condat, 2016]{Condat2016FastBall}
Condat, L. (2016).
\newblock {Fast projection onto the simplex and the l1 ball}.
\newblock {\em Mathematical Programming}, 158(1-2):575--585.

\bibitem[Dai et~al., 2022]{Dai2022DisPFL:Training}
Dai, R., Shen, L., He, F., Tian, X., and Tao, D. (2022).
\newblock {DisPFL: Towards Communication-Efficient Personalized Federated Learning via Decentralized Sparse Training}.

\bibitem[Darlow et~al., 2018]{Darlow2018CINIC-10CIFAR-10}
Darlow, L.~N., Crowley, E.~J., Antoniou, A., and Storkey, A.~J. (2018).
\newblock {CINIC-10 is not ImageNet or CIFAR-10}.

\bibitem[Deng et~al., 2010]{Deng2010ImageNet:Database}
Deng, J., Dong, W., Socher, R., Li, L.-J., {Kai Li}, and {Li Fei-Fei} (2010).
\newblock {ImageNet: A large-scale hierarchical image database}.

\bibitem[Dinh et~al., 2020]{Dinh2020PersonalizedEnvelopes}
Dinh, C.~T., Tran, N.~H., and Nguyen, T.~D. (2020).
\newblock {Personalized Federated Learning with Moreau Envelopes}.
\newblock {\em Advances in Neural Information Processing Systems}, 2020-December.

\bibitem[Dong et~al., 2023]{Dong2023WeiAvg:Diversity}
Dong, F., Abbasi, A., Drew, S., Leung, H., Wang, X., and Zhou, J. (2023).
\newblock {WeiAvg: Federated Learning Model Aggregation Promoting Data Diversity}.

\bibitem[Fellus et~al., 2013]{Fellus2013DecentralizedDatasets}
Fellus, J., Picard, D., and Gosselin, P.~H. (2013).
\newblock {Decentralized K-means using randomized gossip protocols for clustering large datasets}.
\newblock {\em Proceedings - IEEE 13th International Conference on Data Mining Workshops, ICDMW 2013}, pages 599--606.

\bibitem[Gao et~al., 2022]{gao2022survey}
Gao, D., Yao, X., and Yang, Q. (2022).
\newblock A survey on heterogeneous federated learning.

\bibitem[Graf et~al., 2021]{graf2021dissecting}
Graf, F., Hofer, C., Niethammer, M., and Kwitt, R. (2021).
\newblock Dissecting supervised contrastive learning.
\newblock In {\em International Conference on Machine Learning}, pages 3821--3830. PMLR.

\bibitem[He et~al., 2020]{He2020GroupEdge}
He, C., Annavaram, M., and Avestimehr, S. (2020).
\newblock {Group Knowledge Transfer: Federated Learning of Large CNNs at the Edge}.
\newblock {\em Advances in Neural Information Processing Systems}, 33:14068--14080.

\bibitem[He et~al., 2015]{He2015DeepRecognition}
He, K., Zhang, X., Ren, S., and Sun, J. (2015).
\newblock {Deep Residual Learning for Image Recognition}.
\newblock {\em Proceedings of the IEEE Computer Society Conference on Computer Vision and Pattern Recognition}, 2016-December:770--778.

\bibitem[Hendrikx, 2023]{Hendrikx2023AAlgorithms}
Hendrikx, H. (2023).
\newblock {A principled framework for the design and analysis of token algorithms}.
\newblock In {\em Proceedings of The 26th International Conference on Artificial Intelligence and Statistics}, pages 470--489.

\bibitem[Hendrikx et~al., 2020]{Hendrikx2020Dual-FreeReduction}
Hendrikx, H., Bach, F., and Massouli{\'{e}}, L. (2020).
\newblock {Dual-Free Stochastic Decentralized Optimization with Variance Reduction}.
\newblock {\em Advances in Neural Information Processing Systems}, 33:19455--19466.

\bibitem[Hsieh et~al., 2019]{Hsieh2019TheLearning}
Hsieh, K., Phanishayee, A., Mutlu, O., and Gibbons, P.~B. (2019).
\newblock {The Non-IID Data Quagmire of Decentralized Machine Learning}.
\newblock {\em 37th International Conference on Machine Learning, ICML 2020}, PartF168147-6:4337--4348.

\bibitem[Jamali-Rad et~al., 2022]{Jamali-Rad2022FederatedData}
Jamali-Rad, H., Abdizadeh, M., and Singh, A. (2022).
\newblock {Federated Learning With Taskonomy for Non-IID Data}.
\newblock {\em IEEE Transactions on Neural Networks and Learning Systems}.

\bibitem[Jang et~al., 2022]{Jang2022FedClassAvg:Networks}
Jang, J., Ha, H., Jung, D., and Yoon, S. (2022).
\newblock {FedClassAvg: Local Representation Learning for Personalized Federated Learning on Heterogeneous Neural Networks}.
\newblock In {\em Proceedings of the 51st International Conference on Parallel Processing}, pages 1--10.

\bibitem[Jeong and Kountouris, 2023]{Jeong2023PersonalizedDistillation}
Jeong, E. and Kountouris, M. (2023).
\newblock {Personalized Decentralized Federated Learning with Knowledge Distillation}.

\bibitem[Jiang et~al., 2017]{Jiang2017CollaborativeNetworks}
Jiang, Z., Balu, A., Hegde, C., and Sarkar, S. (2017).
\newblock {Collaborative Deep Learning in Fixed Topology Networks}.
\newblock {\em Advances in Neural Information Processing Systems}, 2017-December:5905--5915.

\bibitem[Kairouz et~al., 2019]{Kairouz2019AdvancesLearning}
Kairouz, P., McMahan, H.~B., Avent, B., Bellet, A., Bennis, M., Bhagoji, A.~N., Bonawitz, K., Charles, Z., Cormode, G., Cummings, R., D'Oliveira, R. G.~L., Eichner, H., Rouayheb, S.~E., Evans, D., Gardner, J., Garrett, Z., Gasc{\'{o}}n, A., Ghazi, B., Gibbons, P.~B., Gruteser, M., Harchaoui, Z., He, C., He, L., Huo, Z., Hutchinson, B., Hsu, J., Jaggi, M., Javidi, T., Joshi, G., Khodak, M., Kone{\v{c}}n{\'{y}}, J., Korolova, A., Koushanfar, F., Koyejo, S., Lepoint, T., Liu, Y., Mittal, P., Mohri, M., Nock, R., {\"{O}}zg{\"{u}}r, A., Pagh, R., Raykova, M., Qi, H., Ramage, D., Raskar, R., Song, D., Song, W., Stich, S.~U., Sun, Z., Suresh, A.~T., Tram{\`{e}}r, F., Vepakomma, P., Wang, J., Xiong, L., Xu, Z., Yang, Q., Yu, F.~X., Yu, H., and Zhao, S. (2019).
\newblock {Advances and Open Problems in Federated Learning}.
\newblock {\em Foundations and Trends{\textregistered} in Machine Learning}, 14(1–2):1--210.

\bibitem[Kalofolias, 2016]{Kalofolias2016HowSignals}
Kalofolias, V. (2016).
\newblock {How to Learn a Graph from Smooth Signals}.
\newblock In {\em Proceedings of the 19th International Conference on Artificial Intelligence and Statistics}, pages 920--929. PMLR.

\bibitem[Kalra et~al., 2023]{Kalra2023DecentralizedSharing}
Kalra, S., Wen, J., Cresswell, J.~C., Volkovs, M., and Tizhoosh, H.~R. (2023).
\newblock {Decentralized federated learning through proxy model sharing}.
\newblock {\em Nature Communications 2023 14:1}, 14(1):1--10.

\bibitem[Karimireddy et~al., 2019]{Karimireddy2019SCAFFOLD:Learning}
Karimireddy, S.~P., Kale, S., Mohri, M., Reddi, S.~J., Stich, S.~U., and Suresh, A.~T. (2019).
\newblock {SCAFFOLD: Stochastic Controlled Averaging for Federated Learning}.
\newblock {\em 37th International Conference on Machine Learning, ICML 2020}, PartF168147-7:5088--5099.

\bibitem[Khalil et~al., 2024]{Khalil2024DFML:Learning}
Khalil, Y.~H., Estiri, A.~H., Beitollahi, M., Asadi, N., Hemati, S., Li, X., Zhang, G., and Chen, X. (2024).
\newblock {DFML: Decentralized Federated Mutual Learning}.

\bibitem[Khosla et~al., 2020]{Khosla2020SupervisedLearning}
Khosla, P., Teterwak, P., Wang, C., Sarna, A., Research, G., Tian, Y., Isola, P., Maschinot, A., Liu, C., and Krishnan, D. (2020).
\newblock {Supervised Contrastive Learning}.
\newblock {\em Advances in neural information processing systems}, 33:18661--18673.

\bibitem[Krizhevsky and Hinton, 2009]{Krizhevsky2009LearningImages}
Krizhevsky, A. and Hinton, G. (2009).
\newblock {Learning multiple layers of features from tiny images}.

\bibitem[Krizhevsky et~al., 2012]{Krizhevsky2012ImageNetNetworks}
Krizhevsky, A., Sutskever, I., and Hinton, G.~E. (2012).
\newblock {ImageNet Classification with Deep Convolutional Neural Networks}.
\newblock {\em Advances in Neural Information Processing Systems}, 25.

\bibitem[Li et~al., 2024]{Li2024Prototype-BasedSystems}
Li, B., Gao, W., Xie, J., Gong, M., Wang, L., and Li, H. (2024).
\newblock {Prototype-Based Decentralized Federated Learning for the Heterogeneous Time-Varying IoT Systems}.
\newblock {\em IEEE Internet of Things Journal}, 11(4):6916--6927.

\bibitem[Li et~al., 2021a]{Li2021Model-ContrastiveLearning}
Li, Q., He, B., and Song, D. (2021a).
\newblock {Model-Contrastive Federated Learning}.
\newblock {\em Proceedings of the IEEE Computer Society Conference on Computer Vision and Pattern Recognition}, pages 10708--10717.

\bibitem[Li et~al., 2022a]{Li2022LearningModels}
Li, S., Zhou, T., Tian, X., and Tao, D. (2022a).
\newblock {Learning To Collaborate in Decentralized Learning of Personalized Models}.

\bibitem[Li et~al., 2021b]{Li2021Ditto:Personalization}
Li, T., Hu, S., Beirami, A., and Smith, V. (2021b).
\newblock {Ditto: Fair and Robust Federated Learning Through Personalization}.

\bibitem[Li et~al., 2018]{Li2018FederatedNetworks}
Li, T., Sahu, A.~K., Zaheer, M., Sanjabi, M., Talwalkar, A., and Smith, V. (2018).
\newblock {Federated Optimization in Heterogeneous Networks}.
\newblock {\em Proceedings of Machine learning and systems}, 2:429--450.

\bibitem[Li et~al., 2021c]{li2021fedh2l}
Li, Y., Zhou, W., Wang, H., Mi, H., and Hospedales, T.~M. (2021c).
\newblock Fedh2l: Federated learning with model and statistical heterogeneity.
\newblock {\em arXiv preprint arXiv:2101.11296}.

\bibitem[Li et~al., 2022b]{Li2022TowardsMatching}
Li, Z., Lu, J., Luo, S., Zhu, D., Shao, Y., Li, Y., Zhang, Z., Wang, Y., and Wu, C. (2022b).
\newblock {Towards Effective Clustered Federated Learning: A Peer-to-peer Framework with Adaptive Neighbor Matching}.
\newblock {\em IEEE Transactions on Big Data}, pages 1--16.

\bibitem[{Li Deng}, 2012]{LiDeng2012TheWeb}
{Li Deng} (2012).
\newblock {The MNIST Database of Handwritten Digit Images for Machine Learning Research [Best of the Web]}.
\newblock {\em IEEE Signal Processing Magazine}, 29(6):141--142.

\bibitem[Lian et~al., 2017]{Lian2017CanDescent}
Lian, X., Zhang, C., Zhang, H., Hsieh, C.-J., Zhang, W., and Liu, J. (2017).
\newblock {Can Decentralized Algorithms Outperform Centralized Algorithms? A Case Study for Decentralized Parallel Stochastic Gradient Descent}.
\newblock {\em Advances in Neural Information Processing Systems}, 30.

\bibitem[Lin et~al., 2020]{Lin2020EnsembleLearning}
Lin, T., Kong, L., Stich, S.~U., and Jaggi, M. (2020).
\newblock {Ensemble Distillation for Robust Model Fusion in Federated Learning}.
\newblock {\em Advances in Neural Information Processing Systems}, 2020-December.

\bibitem[Liu et~al., 2022]{Liu2022CompletelyLearning}
Liu, C., Yang, Y., Cai, X., Ding, Y., and Lu, H. (2022).
\newblock {Completely Heterogeneous Federated Learning}.

\bibitem[McMahan et~al., 2017]{McMahan2017Communication-EfficientData}
McMahan, B., Moore, E., Ramage, D., Hampson, S., and Arcas, B. A.~y. (2017).
\newblock {Communication-Efficient Learning of Deep Networks from Decentralized Data}.

\bibitem[Mu et~al., 2021]{Mu2021FedProc:Data}
Mu, X., Shen, Y., Cheng, K., Geng, X., Fu, J., Zhang, T., and Zhang, Z. (2021).
\newblock {FedProc: Prototypical Contrastive Federated Learning on Non-IID data}.
\newblock {\em Future Generation Computer Systems}, 143:93--104.

\bibitem[Netzer et~al., 2011]{Netzer2011ReadingLearning}
Netzer, Y., Wang, T., Coates, A., Bissacco, A., Wu, B., and Ng, A.~Y. (2011).
\newblock {Reading Digits in Natural Images with Unsupervised Feature Learning}.

\bibitem[Oord et~al., 2018]{Oord2018RepresentationCoding}
Oord, A. v.~d., Li, Y., and Vinyals, O. (2018).
\newblock {Representation Learning with Contrastive Predictive Coding}.
\newblock {\em arXiv preprint arXiv:1807.03748}.

\bibitem[Paszke et~al., 2019]{Paszke2019PyTorch:Library}
Paszke, A., Gross, S., Massa, F., Lerer, A., Bradbury, J., Chanan, G., Killeen, T., Lin, Z., Gimelshein, N., Antiga, L., Desmaison, A., K{\"{o}}pf, A., Yang, E., DeVito, Z., Raison, M., Tejani, A., Chilamkurthy, S., Steiner, B., Fang, L., Bai, J., and Chintala, S. (2019).
\newblock {PyTorch: An Imperative Style, High-Performance Deep Learning Library}.
\newblock {\em Advances in Neural Information Processing Systems}, 32.

\bibitem[Pillutla et~al., 2022]{Pillutla2022FederatedPersonalization}
Pillutla, K., Malik, K., Mohamed, A., Rabbat, M., Sanjabi, M., and Xiao, L. (2022).
\newblock {Federated Learning with Partial Model Personalization}.

\bibitem[Regatti et~al., 2022]{Regatti2022ConditionalLearning}
Regatti, J.~R., Lu, S., Gupta, A., and Shroff, N. (2022).
\newblock {Conditional Moment Alignment for Improved Generalization in Federated Learning}.
\newblock In {\em Workshop on Federated Learning: Recent Advances and New Challenges (in Conjunction with NeurIPS 2022)}.

\bibitem[Sattler et~al., 2019]{Sattler2019ClusteredConstraints}
Sattler, F., Muller, K.~R., and Samek, W. (2019).
\newblock {Clustered Federated Learning: Model-Agnostic Distributed Multi-Task Optimization under Privacy Constraints}.
\newblock {\em IEEE Transactions on Neural Networks and Learning Systems}, 32(8):3710--3722.

\bibitem[Shamsian et~al., 2021]{Shamsian2021PersonalizedHypernetworks}
Shamsian, A., Navon, A., Fetaya, E., and Chechik, G. (2021).
\newblock {Personalized Federated Learning using Hypernetworks}.
\newblock {\em Proceedings of Machine Learning Research}, 139:9489--9502.

\bibitem[Shi et~al., 2023a]{Shi2023TowardsTraining}
Shi, Y., Liu, Y., Sun, Y., Lin, Z., Shen, L., Wang, X., and Tao, D. (2023a).
\newblock {Towards More Suitable Personalization in Federated Learning via Decentralized Partial Model Training}.

\bibitem[Shi et~al., 2023b]{Shi2023ImprovingApproaches}
Shi, Y., Shen, L., Wei, K., Sun, Y., Yuan, B., Wang, X., and Tao, D. (2023b).
\newblock {Improving Model Consistency of Decentralized Federated Learning via Sharpness Aware Minimization and Multiple Gossip Approaches}.

\bibitem[Shi et~al., 2023c]{Shi2023ImprovingLearning}
Shi, Y., Shen, L., Wei, K., Sun, Y., Yuan, B., Wang, X., and Tao, D. (2023c).
\newblock {Improving the Model Consistency of Decentralized Federated Learning}.
\newblock {\em Proceedings of Machine Learning Research}, 202:31269--31291.

\bibitem[Soliman et~al., 2020]{Soliman2020DecentralizedCounters}
Soliman, A., Girdzijauskas, S., Bouguelia, M.~R., Pashami, S., and Nowaczyk, S. (2020).
\newblock {Decentralized and Adaptive K-Means Clustering for Non-IID Data Using HyperLogLog Counters}.
\newblock {\em Lecture Notes in Computer Science (including subseries Lecture Notes in Artificial Intelligence and Lecture Notes in Bioinformatics)}, 12084 LNAI:343--355.

\bibitem[Sun et~al., 2022]{Sun2022DecentralizedAveraging}
Sun, T., Li, D., and Wang, B. (2022).
\newblock {Decentralized Federated Averaging}.
\newblock {\em IEEE Transactions on Pattern Analysis and Machine Intelligence}.

\bibitem[Szegedy et~al., 2014]{Szegedy2014GoingConvolutions}
Szegedy, C., Liu, W., Jia, Y., Sermanet, P., Reed, S., Anguelov, D., Erhan, D., Vanhoucke, V., and Rabinovich, A. (2014).
\newblock {Going Deeper with Convolutions}.
\newblock {\em Proceedings of the IEEE Computer Society Conference on Computer Vision and Pattern Recognition}, 07-12-June-2015:1--9.

\bibitem[Tan et~al., 2022]{Tan2022TowardLearning}
Tan, A.~Z., Yu, H., Cui, L., and Yang, Q. (2022).
\newblock {Toward Personalized Federated Learning}.
\newblock {\em IEEE Transactions on Neural Networks and Learning Systems}.

\bibitem[Tan et~al., 2023]{Tan2023PFedSim:Learning}
Tan, J., Zhou, Y., Liu, G., Wang, J.~H., and Yu, S. (2023).
\newblock {pFedSim: Similarity-Aware Model Aggregation Towards Personalized Federated Learning}.

\bibitem[Tan et~al., 2021]{Tan2021FedProto:Clients}
Tan, Y., Long, G., Liu, L., Zhou, T., Lu, Q., Jiang, J., and Zhang, C. (2021).
\newblock {FedProto: Federated Prototype Learning across Heterogeneous Clients}.
\newblock {\em Proceedings of the AAAI Conference on Artificial Intelligence}, 36(8):8432--8440.

\bibitem[Vanhaesebrouck et~al., 2016]{Vanhaesebrouck2016DecentralizedNetworks}
Vanhaesebrouck, P., Bellet, A., and Tommasi, M. (2016).
\newblock {Decentralized Collaborative Learning of Personalized Models over Networks}.
\newblock {\em Proceedings of the 20th International Conference on Artificial Intelligence and Statistics, AISTATS 2017}.

\bibitem[Verbraeken et~al., 2020]{Verbraeken2020ALearning}
Verbraeken, J., Wolting, M., Katzy, J., Kloppenburg, J., Verbelen, T., and Rellermeyer, J.~S. (2020).
\newblock {A Survey on Distributed Machine Learning}.
\newblock {\em ACM Computing Surveys (CSUR)}, 53(2).

\bibitem[Vogels et~al., 2022]{Vogels2022BeyondLearning}
Vogels, T., Hendrikx, H., and Jaggi, M. (2022).
\newblock {Beyond spectral gap: The role of the topology in decentralized learning}.
\newblock In {\em Advances in Neural Information Processing Systems 35 (NeurIPS 2022)}.

\bibitem[Wang et~al., 2023]{Wang2023DoesSupervision}
Wang, L., Zhang, K., Li, Y., Tian, Y., and Tedrake, R. (2023).
\newblock {Does Learning from Decentralized Non-IID Unlabeled Data Benefit from Self Supervision?}
\newblock In {\em 11th International Conference on Learning Representations}.

\bibitem[Wen et~al., 2022]{Wen2022AApplications}
Wen, J., Zhang, Z., Lan, Y., Cui, Z., Cai, J., and Zhang, W. (2022).
\newblock {A survey on federated learning: challenges and applications}.
\newblock {\em International Journal of Machine Learning and Cybernetics 2022 14:2}, 14(2):513--535.

\bibitem[Wu et~al., 2018]{Wu2018UnsupervisedDiscrimination}
Wu, Z., Xiong, Y., Yu, S.~X., and Lin, D. (2018).
\newblock {Unsupervised Feature Learning via Non-Parametric Instance-level Discrimination}.
\newblock {\em Proceedings of the IEEE Computer Society Conference on Computer Vision and Pattern Recognition}, pages 3733--3742.

\bibitem[Xiao et~al., 2017]{Xiao2017Fashion-MNIST:Algorithms}
Xiao, H., Rasul, K., and Vollgraf, R. (2017).
\newblock {Fashion-MNIST: a Novel Image Dataset for Benchmarking Machine Learning Algorithms}.

\bibitem[Xiao et~al., 2007]{Xiao2007DistributedDeviation}
Xiao, L., Boyd, S., and Kim, S.~J. (2007).
\newblock {Distributed average consensus with least-mean-square deviation}.
\newblock {\em Journal of Parallel and Distributed Computing}, 67(1):33--46.

\bibitem[Xu et~al., 2023]{Xu2023PersonalizedCollaboration}
Xu, J., Tong, X., and Huang, S.-L. (2023).
\newblock {Personalized Federated Learning with Feature Alignment and Classifier Collaboration}.
\newblock {\em 11th International Conference on Learning Representations, ICLR 2023 - Conference Track Proceedings}.

\bibitem[Ye et~al., 2023]{Ye2023HeterogeneousChallenges}
Ye, M., Fang, X., Du, B., Yuen, P.~C., and Tao, D. (2023).
\newblock {Heterogeneous Federated Learning: State-of-the-art and Research Challenges}.
\newblock {\em ACM Computing Surveys}, 56(3):1--44.

\bibitem[Zantedeschi et~al., 2020]{Zantedeschi2020FullyGraphs}
Zantedeschi, V., Bellet, A., and Tommasi, M. (2020).
\newblock {Fully Decentralized Joint Learning of Personalized Models and Collaboration Graphs}.

\bibitem[Zhang et~al., 2021]{Zhang2021ParameterizedLearning}
Zhang, J., Guo, S., Ma, X., Wang, H., Xu, W., and Wu, F. (2021).
\newblock {Parameterized Knowledge Transfer for Personalized Federated Learning}.
\newblock {\em Advances in Neural Information Processing Systems}, 13:10092--10104.

\bibitem[Zhang et~al., 2024]{zhang2024energy}
Zhang, X., Chiu, C.-C., and He, T. (2024).
\newblock Energy-efficient decentralized learning via graph sparsification.
\newblock {\em arXiv preprint arXiv:2401.03083}.

\bibitem[Zhang et~al., 2017]{Zhang2017ShuffleNet:Devices}
Zhang, X., Zhou, X., Lin, M., and Sun, J. (2017).
\newblock {ShuffleNet: An Extremely Efficient Convolutional Neural Network for Mobile Devices}.
\newblock {\em Proceedings of the IEEE Computer Society Conference on Computer Vision and Pattern Recognition}, pages 6848--6856.

\bibitem[Zhu et~al., 2021]{Zhu2021Data-FreeLearning}
Zhu, Z., Hong, J., and Zhou, J. (2021).
\newblock {Data-Free Knowledge Distillation for Heterogeneous Federated Learning}.
\newblock {\em Proceedings of Machine Learning Research}, 139:12878--12889.

\end{thebibliography}
\bibliographystyle{apalike}
%%%%%%%%%%%%%%%%%%%%%%%%%%%%%%%%
% APPENDIX
%%%%%%%%%%%%%%%%%%%%%%%%%%%%%%%
\newpage

\appendix

\section{Implementation Details}
\label{sec:appendix-impl}

In this section, we elaborate on the implementation and training details of \ourmethod{}.

\subsection{Architecture Details}
\label{sec:appendix-arch}

\ourmethod{} is implemented using the PyTorch framework \citep{Paszke2019PyTorch:Library}. In the case of heterogeneous model settings, we select a backbone network ($f_{\theta}$) randomly from a set of four options: GoogLeNet \citep{Szegedy2014GoingConvolutions}, ShuffleNet \citep{Zhang2017ShuffleNet:Devices}, ResNet18 \citep{He2015DeepRecognition}, and AlexNet \citep{Krizhevsky2012ImageNetNetworks}. Conversely, for homogeneous model settings, all clients are equipped with a ResNet18 backbone. The projection ($f_{\psi}$) and prediction ($g_{\phi}$) heads consist of 2- and 1-layer MLPs, respectively. Batch Normalization (BN) and ReLU activation are applied solely to the hidden layer of the projection network. Meanwhile, for the prediction head, ReLU activation is on the output layer. We use a resolution of $32 \times 32$ for the input images of CIFAR$10$, CINIC$10$ and SVHN. While for MNIST and FashionMNIST, the image resolution is $28 \times 28$. Unless otherwise specified, all experiments and models use a latent embedding dimension of $d_z = 512$. The number of prototypes used in the experiment corresponds to the number of globally available classes $K=10$ per dataset. It is assumed that ground truth mapping of the samples to the class labels is constant across all the clients. The prototypes are implemented as a linear layer with input dimension $d_p = 512$ and the output dimension $K=10$.

\subsection{Training Details}
\label{sec:appendix-train}

\ourmethod{} is trained on the training splits of CINIC$10$, SVHN, CIFAR$10$, MNIST and FashionMNIST using a batch size of $64$ images for all the datasets. We use the Adam optimizer with a learning rate of $0.0001$, weight decay $0$ and momentum $0.5$. The temperature scalar for the supervised contrastive loss is set to $\tau = 100$. For the graph learning object, the hyper-parameter $\mu_1$ and $\mu_2$ are empirically set as $0.5$ and $0.1$ respectively. In the graph regularization term, the parameter $\beta > 0$ allows us to control the sparsity of the network. For our experiments, the value of $\beta$ is fixed at $0.5$.

Note that at the beginning of the training, the classifier weights are unreliable for inferring the task similarity. Thus, we allow for a warmup phase $T_{\textup{thr}}$ (empirically $T_{\textup{thr}} = 100$ epochs) during which the graph learning objective is not used. However, we use the prototype for inter-client alignment of latent representations from the start of the training. Overall, we train \ourmethod{} for $T=400$ epochs with $E=1$ local iteration for each client at every round.

\subsection{Image Augmentations}
\label{sec:appendix-augm}

The set of image augmentations $\mathcal{A}$ that were applied in \ourmethod{} is shown in Tab.~\ref{tab:data-augmentations}. These augmentations were applied to the input images for all training datasets. It follows a common data augmentation strategy in SSL, including RandomResizedCrop (with scale in $[0.2, 1.0]$), random ColorJitter \citep{Wu2018UnsupervisedDiscrimination} of \{brightness, contrast, saturation, hue\} with a probability of $0.8$, RandomGrayScale with a probability of $0.2$, random GaussianBlur with a probability of $0.5$ and a Gaussian kernel in $\left[0.1, 2.0\right]$, and finally, RandomHorizontalFlip with a probability of $0.5$. Following \citet{Jang2022FedClassAvg:Networks}, we followed this augmentation strategy for all training images before being passed to the backbone.

\begin{table}[ht]
\centering
\renewcommand{\arraystretch}{1.2}
\caption{\small Pytorch-like descriptions of the data augmentation profiles applied in \ourmethod{}.}
% % \vspace{-0.1cm}
\label{tab:data-augmentations}
\resizebox{0.7\textwidth}{!}{%
\begin{tabular}{@{}l@{}}
\toprule
\multicolumn{1}{c}{\textbf{Description}} \\ \midrule \midrule
RandomResizedCrop(size=$224$, scale=$(0.2, 1)$)                                        \\
RandomApply([ColorJitter(brightness=$0.4$, contrast=$0.4$, saturation=$0.4$, hue=$0.1$)], p=$0.8$)                                        \\
RandomGrayScale(p=$0.2$)                                         \\
RandomApply([GaussianBlur($[0.1, 2.0]$)], p=$0.5$)                                        \\
RandomHorizontalFlip(p=$0.5$)                                        \\ \bottomrule
\end{tabular}%
}
\end{table}

\clearpage

\section{Extended Results}
\label{sec:extended_res}

\begin{table}[ht]
\caption{Average test accuracies in (\% ± std) for \emph{homogeneous} models. Style: \textbf{best} and \underline{second best}.}
\label{tab:res_homo}
  \centering
    \aboverulesep = 0pt
    \belowrulesep = 0pt
  \renewcommand{\arraystretch}{1.4}
  \resizebox{\linewidth}{!}{
  \begin{tabular}{l|c|c|c|c|c|c|c|c}
    \toprule
    \textbf{Method}& \textbf{Model Agnostic} & \textbf{Data Heterogeneity} & \textbf{Setting} & \cellcolor{tabPurple2}\textbf{CINIC$10$} & \cellcolor{tabPurple}\textbf{SVHN} & \cellcolor{tabOrange}\textbf{CIFAR$10$} & \cellcolor{tabBlue2}\textbf{MNIST} & \cellcolor{tabGreen}\textbf{FashionMNIST}\\
    % \noalign{\vskip 2pt}
    \midrule
    %\midrule
    \textbf{Local} & - & \texttt{Sc. 1} & \texttt{Local} & 45.00 ($\pm$08.18) & 58.93 ($\pm$13.99) & 55.67 ($\pm$11.29) & 81.80 ($\pm$12.37) & 74.60 ($\pm$12.11) \\
    \midrule
    \textbf{FedSim} \citep{Pillutla2022FederatedPersonalization} & \checkmark & \texttt{Sc. 1} & \texttt{Cent.} & 64.80 ($\pm$05.79) & 72.60 ($\pm$05.47) & 69.13 ($\pm$07.84) & 97.20 ($\pm$02.52) & 91.07 ($\pm$05.30) \\
    \textbf{FedClassAvg} \citep{Jang2022FedClassAvg:Networks} & \checkmark & \texttt{Sc. 1} & \texttt{Cent.} & \underline{70.80 ($\pm$06.50)} & \underline{80.00 ($\pm$06.24)} & \underline{75.60 ($\pm$07.08)} & 97.00 ($\pm$02.59) & \textbf{91.53 ($\pm$04.41)} \\
    \textbf{FedProto} \citep{Tan2021FedProto:Clients} & \checkmark & \texttt{Sc. 1} & \texttt{Cent.} & 66.07 ($\pm$05.01) & 79.80 ($\pm$04.84) & 72.00 ($\pm$07.63) & \textbf{98.26 ($\pm$01.74)} & 91.20 ($\pm$05.01) \\
    \noalign{\vskip -2pt}
    \hdashline
    \rowcolor{tabBlue}
    \textbf{\ourmethod{} (Ours)} & \checkmark & \texttt{Sc. 1} & \texttt{P2P} & \textbf{72.60 ($\pm$05.21)} & \textbf{82.40 ($\pm$04.21)} & \textbf{76.27 ($\pm$08.97)} & \underline{97.93 ($\pm$01.86)} & \underline{91.27 ($\pm$04.51)}\\
    \midrule
    \textbf{FedAvg}  \citep{McMahan2017Communication-EfficientData} & - &  \texttt{Sc. 1} & \texttt{Cent.} & \underline{79.20 ($\pm$04.85)} & 89.80 ($\pm$03.68) & 81.00 ($\pm$05.75) & \underline{99.07 ($\pm$01.04)} & 93.93 ($\pm$03.31)\\
    \textbf{FLT} \citep{Jamali-Rad2022FederatedData} & - & \texttt{Sc. 1} & \texttt{Cent.} & 77.20 ($\pm$06.35) & \underline{88.13 ($\pm$03.97)} & \underline{82.20 ($\pm$05.06)} & 98.67 ($\pm$01.63) & \underline{93.73 ($\pm$03.32)} \\
    \textbf{pFedSim} \citep{Tan2023PFedSim:Learning} & - & \texttt{Sc. 1} & \texttt{Cent.} & 77.07 ($\pm$05.65) & 88.73 ($\pm$03.76) & 79.86 ($\pm$05.81) & 99.00 ($\pm01.57$) & 92.80 ($\pm$03.68) \\
    %\midrule
    \textbf{DFedAvgM} \citep{Sun2022DecentralizedAveraging} & - & \texttt{Sc. 1} & \texttt{P2P} & 78.60 ($\pm$05.49) & \underline{90.40 ($\pm$03.54)} & 81.26 ($\pm$06.63) & 98.93 ($\pm$01.24) & 93.47 ($\pm$03.98)  \\
   \noalign{\vskip -2pt}
    \hdashline
    \rowcolor{tabBlue} \textbf{\ourmethod{}+ (Ours)} & - & \texttt{Sc. 1} & \texttt{P2P} & \textbf{83.40 ($\pm$02.91)} & \textbf{91.67 ($\pm$03.93)} & \textbf{82.93 ($\pm$04.96)} & \textbf{99.13 ($\pm$01.42)} & \textbf{94.40 ($\pm$03.47)} \\
    \midrule  \midrule
    \textbf{Local} & - & \texttt{Sc. 2} & \texttt{Local} & 34.48 ($\pm$05.33) & 57.00 ($\pm$10.97) & 43.76 ($\pm$07.37) & 77.52 ($\pm$09.42) & 73.95 ($\pm$08.59) \\
    \midrule
    \textbf{FedSim} \citep{Pillutla2022FederatedPersonalization} & \checkmark & \texttt{Sc. 2} & \texttt{Cent.} & 50.19 ($\pm$06.81) & 68.19 ($\pm$03.94) & 59.91 ($\pm$06.78) & 95.09 ($\pm$02.63) & 85.86 ($\pm$04.42) \\
    \textbf{FedClassAvg} \citep{Jang2022FedClassAvg:Networks} & \checkmark & \texttt{Sc. 2} & \texttt{Cent.} & \underline{58.00 ($\pm$05.17)} & \underline{78.05 ($\pm$02.93)} & \underline{64.38 ($\pm$04.38)} & 95.76 ($\pm$02.34) & 85.62 ($\pm$04.02) \\
    \textbf{FedProto} \citep{Tan2021FedProto:Clients} & \checkmark & \texttt{Sc. 2} & \texttt{Cent.} & 51.33 ($\pm$06.62) & 75.33 ($\pm$04.23) & 60.57 ($\pm$05.89) & \textbf{97.00 ($\pm$02.03)} & \textbf{87.00 ($\pm$04.34)} \\ 
    \noalign{\vskip -2pt}
    \hdashline
    \rowcolor{tabBlue}
    \textbf{\ourmethod{} (Ours)} & \checkmark & \texttt{Sc. 2} & \texttt{P2P} & \textbf{59.14 ($\pm$05.66)} & \textbf{78.71 ($\pm$04.56)} & \textbf{64.81 ($\pm$08.47)} & \underline{96.38 ($\pm$01.87)} & \underline{87.24 ($\pm$04.29)}  \\
    \midrule
    \textbf{FedAvg}  \citep{McMahan2017Communication-EfficientData} & - & \texttt{Sc. 2} & \texttt{Cent.} & 67.86 ($\pm$05.46) & 88.24 ($\pm$03.15) & 76.43 ($\pm$06.02) & \underline{97.76 ($\pm$01.17)} & \underline{90.71 ($\pm$03.39)} \\
    \textbf{FLT} \citep{Jamali-Rad2022FederatedData} & - & \texttt{Sc. 2} & \texttt{Cent.} & 65.81 ($\pm$07.73) & \underline{88.29 ($\pm$02.92)} & 75.47 ($\pm$05.06) & 97.71 ($\pm$01.73) & 89.33 ($\pm$04.22) \\
    \textbf{pFedSim} \citep{Tan2023PFedSim:Learning} & - & \texttt{Sc. 2} & \texttt{Cent.} & 68.52 ($\pm$07.53) & 88.33 ($\pm$02.72) & 75.90 ($\pm$04.88) & 97.62 ($\pm$03.02) & 90.14 ($\pm$04.15) \\
    % \midrule
    \textbf{DFedAvgM} \citep{Sun2022DecentralizedAveraging} & - & \texttt{Sc. 2} & \texttt{P2P} & \underline{70.62 ($\pm$07.06)} & 88.14 ($\pm$03.40) & \underline{76.67 ($\pm$04.62)} & 97.76 ($\pm$01.51) & 90.38 ($\pm$03.46) \\
    \noalign{\vskip -2pt}
    \hdashline
    \rowcolor{tabBlue} \textbf{\ourmethod{}+ (Ours)} & - & \texttt{Sc. 2} & \texttt{P2P} & \textbf{72.76 ($\pm$04.63)} & \textbf{89.38 ($\pm$03.28)} & \textbf{77.86 ($\pm$04.33)} & \textbf{97.76 ($\pm$01.09)} & \textbf{91.24 ($\pm$04.58)}  \\
    \midrule \midrule
    %\midrule
    \textbf{Local} & - & \texttt{Sc. 3} & \texttt{Local} & 43.87 ($\pm$08.19) & 59.33 ($\pm$14.72) & 54.47 ($\pm$11.33) & 80.87 ($\pm$13.41) & 73.93 ($\pm$12.48) \\
    \midrule
    \textbf{FedSim} \citep{Pillutla2022FederatedPersonalization} & \checkmark & \texttt{Sc. 3} & \texttt{Cent.} & 60.40 ($\pm$05.37) & 63.00 ($\pm$05.11) & 65.47 ($\pm$10.10) & 95.00 ($\pm$02.92) & 90.00 ($\pm$04.61) \\
    \textbf{FedClassAvg} \citep{Jang2022FedClassAvg:Networks} & \checkmark & \texttt{Sc. 3} & \texttt{Cent.} & \underline{66.73 ($\pm$05.97)} & \underline{74.87 ($\pm$06.06)} & \underline{70.87 ($\pm$08.28)} & 96.27 ($\pm$02.58) & 89.47 ($\pm$04.19) \\
    \textbf{FedProto} \citep{Tan2021FedProto:Clients} & \checkmark & \texttt{Sc. 3} & \texttt{Cent.} & 61.00 ($\pm$05.02) & 72.93 ($\pm$05.96) & 66.27 ($\pm$08.21) & \textbf{97.33 ($\pm$01.89)} & \underline{91.13 ($\pm$04.26)} \\
    \noalign{\vskip -2pt}
    \hdashline
    \rowcolor{tabBlue}
    \textbf{\ourmethod{} (Ours)} & \checkmark & \texttt{Sc. 3} & \texttt{P2P} & \textbf{67.00 ($\pm$06.04)} & \textbf{76.73 ($\pm$05.79)} & \textbf{71.27 ($\pm$08.12)} & \underline{96.53 ($\pm$02.04)} & \textbf{92.00 ($\pm$04.26)} \\
    \midrule
    \textbf{FedAvg}  \citep{McMahan2017Communication-EfficientData} & - &  \texttt{Sc. 3} & \texttt{Cent.} & 74.33 ($\pm$04.48) & 87.53 ($\pm$05.29) & 79.67 ($\pm$05.62) & 98.13 ($\pm$01.81) & 92.80 ($\pm$04.72) \\
    \textbf{FLT} \citep{Jamali-Rad2022FederatedData} & - & \texttt{Sc. 3} & \texttt{Cent.} & 73.53 ($\pm$04.85) & 88.20 ($\pm$04.31) & 79.67 ($\pm$05.64) & 99.00 ($\pm$00.83) & 92.20 ($\pm$05.24) \\
    \textbf{pFedSim} \citep{Tan2023PFedSim:Learning} & - & \texttt{Sc. 3} & \texttt{Cent.} & 73.40 ($\pm$06.10) & 87.60 ($\pm$04.42) & 79.07 ($\pm$06.20) & \textbf{98.53 ($\pm$01.26)} & 92.53 ($\pm$04.68) \\
    %\midrule
    \textbf{DFedAvgM} \citep{Sun2022DecentralizedAveraging} & - & \texttt{Sc. 3} & \texttt{P2P} & \underline{75.87 ($\pm$06.65)} & \underline{89.47 ($\pm$04.69)} & \textbf{81.80 ($\pm$06.01)} & 97.80 ($\pm$05.22) & \underline{93.20 ($\pm$04.60)} \\
    \noalign{\vskip -0.5pt}
    \hdashline
    \rowcolor{tabBlue} \textbf{\ourmethod{}+ (Ours)} & - & \texttt{Sc. 3} & \texttt{P2P} & \textbf{80.00 ($\pm$04.92)} & \textbf{90.40 ($\pm$03.67)} & \underline{80.13 ($\pm$07.10)} & \underline{98.13 ($\pm$02.21)} & \textbf{93.33 ($\pm$03.98)} \\
    \midrule  \midrule
    \textbf{Local} & - & \texttt{Sc. 4} & \texttt{Local} & 35.00 ($\pm$05.85) & 56.81 ($\pm$10.91) & 43.00 ($\pm$08.04) & 76.95 ($\pm$09.40) & 73.86 ($\pm$09.48) \\
    \midrule
    \textbf{FedSim} \citep{Pillutla2022FederatedPersonalization} & \checkmark & \texttt{Sc. 4} & \texttt{Cent.} & 47.29 ($\pm$06.17) & 60.52 ($\pm$05.80) & 55.33 ($\pm$05.76) & 93.62 ($\pm$02.48) & 84.33 ($\pm$04.06) \\
    \textbf{FedClassAvg} \citep{Jang2022FedClassAvg:Networks} & \checkmark & \texttt{Sc. 4} & \texttt{Cent.} & \underline{54.48 ($\pm$04.49)} & \underline{73.00 ($\pm$04.46)} & \underline{59.95 ($\pm$04.98)} & 94.05 ($\pm$03.33) & 84.33 ($\pm$04.19) \\
    \textbf{FedProto} \citep{Tan2021FedProto:Clients} & \checkmark & \texttt{Sc. 4} & \texttt{Cent.} & 47.95 ($\pm$07.24) & 70.05 ($\pm$04.40) & 53.86 ($\pm$06.98) & \textbf{96.05 ($\pm$01.71)} & \textbf{86.48 ($\pm$04.20)} \\ 
    \noalign{\vskip -2pt}
    \hdashline
    \rowcolor{tabBlue}
    \textbf{\ourmethod{} (Ours)} & \checkmark & \texttt{Sc. 4} & \texttt{P2P} & \textbf{55.19 ($\pm$04.94)} & \textbf{73.38 ($\pm$04.40)} & \textbf{61.00 ($\pm$06.20)} & \underline{95.48 ($\pm$02.02)} & \underline{85.62 ($\pm$03.83)} \\
    \midrule
    \textbf{FedAvg}  \citep{McMahan2017Communication-EfficientData} & - & \texttt{Sc. 4} & \texttt{Cent.} & 63.33 ($\pm$05.91) & \underline{87.19 ($\pm$04.13)} & 74.38 ($\pm$04.21) & 96.38 ($\pm$02.57) & 88.62 ($\pm$04.45) \\
    \textbf{FLT} \citep{Jamali-Rad2022FederatedData} & - & \texttt{Sc. 4} & \texttt{Cent.} & 59.90 ($\pm$06.94) & 86.86 ($\pm$03.62) & 73.43 ($\pm$08.54) & \underline{97.52 ($\pm$01.49)} & 88.76 ($\pm$04.18) \\
    \textbf{pFedSim} \citep{Tan2023PFedSim:Learning} & - & \texttt{Sc. 4} & \texttt{Cent.} & \underline{63.95 ($\pm$05.69)} & 85.95 ($\pm$03.71) & 72.67 ($\pm$04.66) & 97.05 ($\pm$01.78) & 89.29 ($\pm$03.72) \\
    % \midrule
    \textbf{DFedAvgM} \citep{Sun2022DecentralizedAveraging} & - & \texttt{Sc. 4} & \texttt{P2P} & 63.57 ($\pm$08.00) & 87.14 ($\pm$03.78) & \underline{76.67 ($\pm$04.18)} & \textbf{97.81 ($\pm$02.09)} & \underline{88.62 ($\pm$04.20)} \\
    \noalign{\vskip -2pt}
    \hdashline
    \rowcolor{tabBlue} \textbf{\ourmethod{}+ (Ours)} & - & \texttt{Sc. 4} & \texttt{P2P} &  \textbf{70.86 ($\pm$04.96)} & \textbf{90.09 ($\pm$03.52)} & \textbf{77.19 ($\pm$05.77)} & 97.19 ($\pm$01.52) & \textbf{89.57 ($\pm$02.77)} \\
    \bottomrule
  \end{tabular}
  }
\end{table}

\textbf{Model homogeneous setting.} In the model homogeneous setting, we additionally compare against \citep{McMahan2017Communication-EfficientData, Jamali-Rad2022FederatedData, Tan2023PFedSim:Learning}. \sayak{In the P2P setting, existing model heterogeneous baselines require additional information in the form of data \citep{li2021fedh2l} or models parameters \citep{Kalra2023DecentralizedSharing, Khalil2024DFML:Learning} for knowledge distillation making it difficult to compare with \ourmethod{}. Therefore, we only compare against the model homogeneous baseline \citep{Sun2022DecentralizedAveraging}.} To enable a fair comparison, here we evaluate \ourmethod{} in two different configurations.  First, akin to the model heterogeneous setting, the backbone model parameters $f_\theta$ are not aggregated. Second, we introduce \ourmethod{}+, which further aggregates the backbones in addition to our proposed method. The result for scenarios $1-4$ is reported in Tab.~\ref{tab:res_homo}. On challenging datasets such as CINIC$10$, SVHN and CIFAR$10$, we observe performance gains of around $1-2\%$ with \ourmethod{} over the model agnostic baselines, while \ourmethod{}+ outperforms the baselines achieving up to a $1-7\%$ improvement.

\begin{figure}[t]
    \centering
    \includegraphics[width=\linewidth]{figures/updates/nol.png}
    \caption{Client data distribution, \emph{learned} collaboration graph (part (b)) and latent embedding (part (c)) with $M=10$ for Scenario $1$.}\vspace{1em}
    \label{fig:col_graph_nol}
    \includegraphics[width=\linewidth]{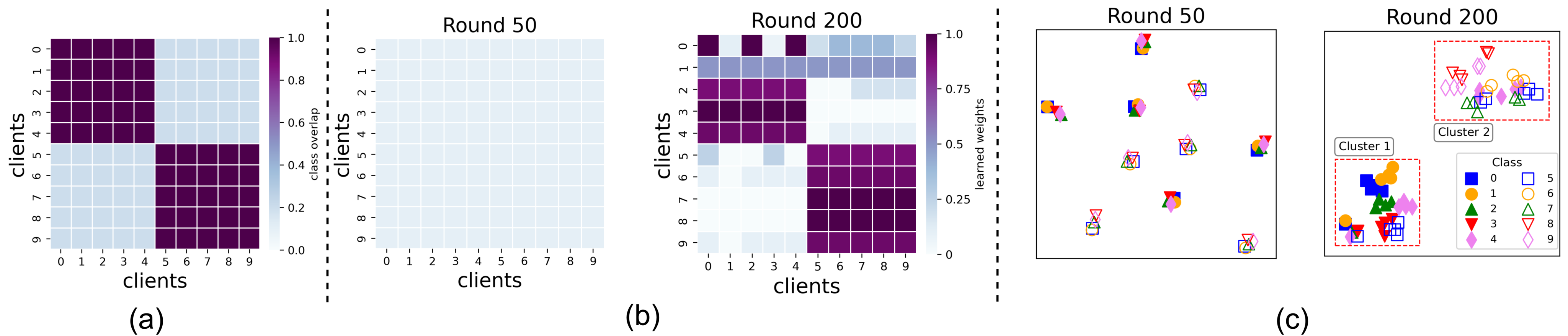}
    \caption{Client data distribution, \emph{learned} collaboration graph (part (b)) and latent embedding (part (c)) with $M=10$ for Scenario $2$.}
    \label{fig:col_graph_ol}
\end{figure}

\textbf{Learned collaboration graph.} We visualize the learned weight matrix $\bm{W}$ and the latent embedding $z$ for both Scenario $1$ and $2$ with $M=10$ clients and $C=2$ clusters. In Scenario $1$, as is shown in Fig.~\ref{fig:col_graph_nol}-(a), clients $0$-$4$ are assigned to cluster $1$ with samples from classes $0$-$4$, while clients $5$-$9$ are assigned to cluster $2$ with samples from classes $5$-$9$. It can be observed in Fig.~\ref{fig:col_graph_nol}-(b) that \ourmethod{} correctly identifies the true clusters as communication rounds progress ($T$:$50$ $\rightarrow$ $200$) assigning equal weights to clients within the same cluster and almost zero weight to clients in the other cluster. It is further reflected in the latent embeddings in Fig.~\ref{fig:col_graph_nol}-(c), where two distinct clusters emerge. For Scenario $2$, as is shown in Fig.~\ref{fig:col_graph_ol}-(a), clients $0$-$4$ are assigned to cluster $1$ with samples from classes $0$-$5$, while clients $5$-$9$ are assigned to cluster $2$ with samples from classes $4$-$9$. Despite the overlapping class assignments, as can be seen in Fig.~\ref{fig:col_graph_ol}-(b), \ourmethod{} successfully identifies the true clusters. Moreover, two clusters of latent embeddings can also be seen in Fig.~\ref{fig:col_graph_ol}-(c) with embeddings of class $4$ and $5$ appearing in both.

\begin{figure}[ht]
    \centering
    \begin{subfigure}{0.45\linewidth}
        \centering
        \includegraphics[width=\linewidth]{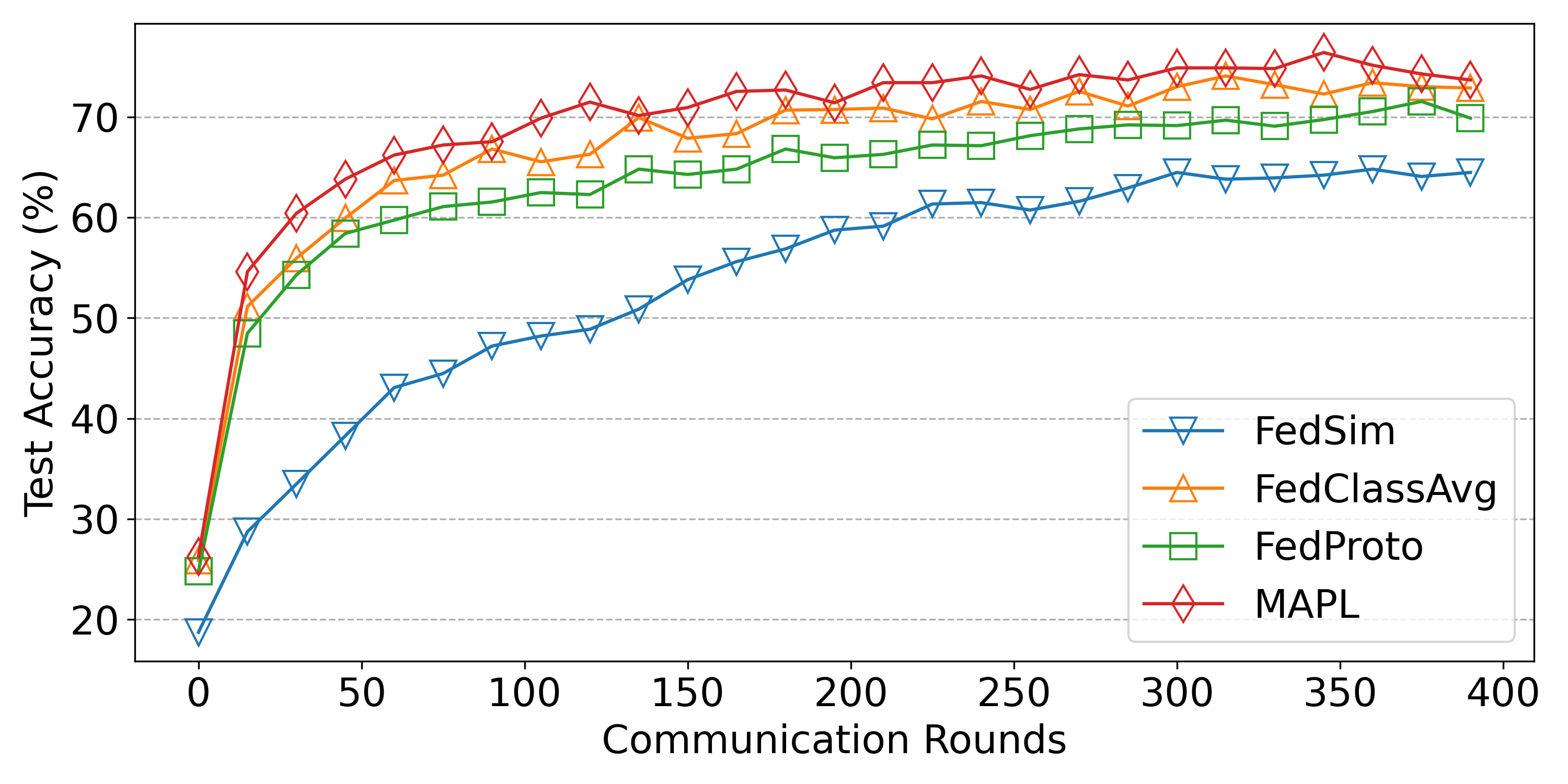}
        % \caption{}
    \end{subfigure}
    \begin{subfigure}{0.45\linewidth}
        \centering
        \includegraphics[width=\linewidth]{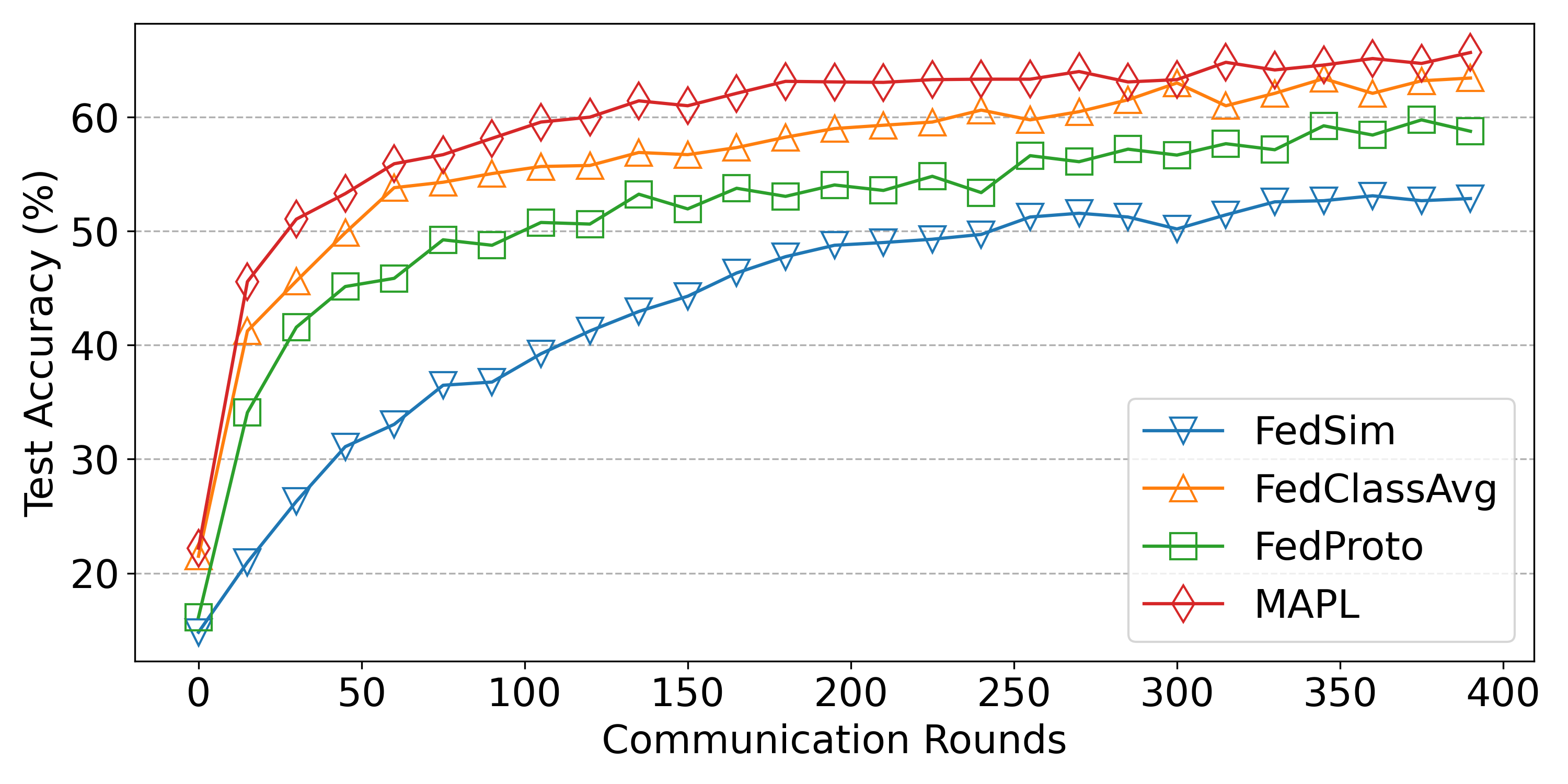}
        % \caption{}
    \end{subfigure}
    \caption{Convergence graph of \emph{test} accuracies for Scenario 1 (left) and Scenario 2 (right), CIFAR$10$, M=$20$.}
    \label{fig:conv}
\end{figure}

\textbf{Convergence.} We demonstrate the convergence graph of \ourmethod{} comparing it against prior art in Fig.~\ref{fig:conv}. As can be seen, in both scenarios, \ourmethod{} converges relatively faster than all other baselines.

\section{Additional Details on Datasets and Baselines}

\subsection{Datasets}
\label{sec:appendix-datasets}

\textbf{CIFAR$10$}. The CIFAR$10$ \citep{Krizhevsky2009LearningImages} dataset is a widely used benchmark in the field of machine learning and computer vision. It consists of $60,000$ $32\times32$ color images in $10$ classes, with $6,000$ images per class. The dataset is split into $50,000$ training images and $10,000$ testing images. The classes are mutually exclusive and represent common objects in everyday life, such as airplanes, automobiles, birds, cats, deer, dogs, frogs, horses, ships, and trucks. 

\textbf{CINIC$10$}. The CINIC$10$ \citep{Darlow2018CINIC-10CIFAR-10} dataset was proposed to bridge the gap between the CIFAR$10$ dataset and a much larger benchmarking dataset ImageNet \citep{Deng2010ImageNet:Database}. It is $4.5\times$ the size of CIFAR$10$ and contains images from both CIFAR$10$ and ImageNet. Further, the distribution shift presents an additional challenge as the images originate from two different datasets. It has a total of $270,000$ images in $10$ classes split into three equal subsets - train, validation, and test - each of which contains $90,000$ images. 

\textbf{MNIST}. The MNIST \citep{LiDeng2012TheWeb} dataset consists of a collection of $28\times28$ pixel grayscale images of handwritten digits ($0$ to $9$). There are $60,000$ training images and 10,000 test images. The dataset is commonly used for training and testing various machine learning models, especially in the context of image recognition and classification tasks. 

\textbf{SVHN}. The Street View House Numbers (SVHN) \citep{Netzer2011ReadingLearning} dataset is a widely used collection of labelled images which is conceptually similar to MNIST in the sense it consists of images of small cropped digits. However, it is diverse, featuring a wide range of fonts, styles, and perspectives, making it a challenging testbed. Developed by Google, it consists of over $600,000$ $32\times32$ RGB images of house numbers extracted from Google Street View.

\textbf{FashionMNIST}. The FashionMNIST \citep{Xiao2017Fashion-MNIST:Algorithms} dataset consists of $28\times28$ grayscale images of fashion items, such as clothing, footwear, and accessories. There are $10$ different categories, each representing a specific type of fashion item, including T-shirts/tops, trousers, pullovers, dresses, coats, sandals, shirts, sneakers, bags, and ankle boots. Similar to the MNIST dataset, FashionMNIST contains $60,000$ training images and $10,000$ test images. It serves as a more challenging replacement for MNIST.

\subsection{Baselines}

\textbf{Local} is the classical way for training personalization. Each client only trains a local model and does not communicate with others. To ensure a fair comparison, we trained the local models for $E=400$ local epochs.

\textbf{FedAvg} \citep{McMahan2017Communication-EfficientData} utilizes a central server to aggregate the locally trained model parameters at the end of the global round. It does not support model heterogeneity, i.e., it requires all the local models to have a similar architecture.

\textbf{DFedAvgM} \citep{Sun2022DecentralizedAveraging} is the decentralized FedAvg with momentum. Unlike FedAvg, it does not consider the presence of a central server. During each global round, the clients locally update the model parameters, followed by the aggregation of the parameters over the neighborhood. 

\textbf{FedSim} \citep{Pillutla2022FederatedPersonalization} considers a decoupled view of the client models, i.e., each client model is comprised of a backbone and a prediction head. During a global round, first, clients locally update the model parameters. Then, only the prediction head is aggregated over all clients with the help of a central server. This setting further allows each client to have a heterogeneous backbone while sharing a common architecture for the prediction head.

\textbf{FLT} \citep{Jamali-Rad2022FederatedData} introduces a method based on FedAvg where instead of aggregating the model parameters over all clients, facilitates collaboration among similar clients by inferring task-relatedness between clients in a one-off process. FLT can flexibly handle generic client relatedness as well as decomposing it into (disjoint) cluster formation. We follow the original implementation\footnote{\href{https://github.com/hjraad/FLT}{https://github.com/hjraad/FLT}} for our experiments.

\textbf{pFedSim} \citep{Tan2023PFedSim:Learning} also attempts to identify related clients similar to FLT. Akin to FedSim and \ourmethod{}, it also considers a decoupled view of the local models. The models are trained locally during each global round, followed by aggregation of the model parameters from the feature extraction backbones. The classifier layer is not aggregated across clients. The authors then use the locally trained classifier heads to infer task-relatedness. 

\textbf{FedProto} \citep{Tan2021FedProto:Clients} proposes a method to collaborate among clients to train heterogeneous models considering the presence of a central server. Instead of explicitly aggregating the model parameters, it uses class-wise feature prototypes to align the classes in the latent space across clients. The prototype for a particular class is computed as an average over all the features pertaining to the samples from the corresponding class in the local dataset. At the end of each global round, the local prototypes are aggregated over all the clients. We adapt the original implementation\footnote{\href{https://github.com/yuetan031/fedproto}{https://github.com/yuetan031/fedproto}} for our study.

\textbf{FedClassAvg} \citep{Jang2022FedClassAvg:Networks} allows collaborate among heterogeneous clients similar to FedProto. Contrary to explicitly aligning the representations in the latent space using class-wise feature representations, FedClassAvg implicitly aligns the clients using the classifier layer. Thus, during each global round, the models are trained locally by each client, followed by the aggregation of the classifier layers across all the clients. Additionally, they use supervised contrastive loss during local training similar to \ourmethod{} to ensure better separability among the classes. We adopted the original implementation\footnote{\href{https://github.com/hukla/FedClassAvg}{https://github.com/hukla/FedClassAvg}}.

\subsection{System Requirements}

For the simulations, each client is trained sequentially during the global rounds, which allowed the experiments to be performed on a single A$40$ NVIDIA GPU. To ensure reproducibility, we fixed a seed value. However, the exact number in the results might still vary between runs in different environments due to the non-deterministic implementations in PyTorch. Nonetheless, the relative performance of the different algorithms should remain the same. We will release the code publicly upon acceptance.

\section{Pseudocode}
\label{sec:pymapl}

This section includes the algorithms for the training methodology of \ourmethod{} in a Pytorch-like pseudocode format. Algorithm~\ref{alg:mapl_pytorch} provides an overview of the complete training methodology of \ourmethod{} and is equivalent to Algorithm~\ref{algo:mapl_overall}. We describe the personalized model learning algorithm of \ourmethod{} in Algorithm~\ref{alg:pml_pytorch}, which is analogous to Algorithm~\ref{algo:mapl_gl} while  Algorithm~\ref{alg:pml_pytorch} demonstrates the collaborative graph learning process also summarized in Algorithm~\ref{alg:cgl_pytorch}.

\begin{algorithm}[ht]
\small
\SetAlgoLined
\DontPrintSemicolon
\SetNoFillComment
        \textcolor{darkgreen}{\# $T$: number of global rounds}\\
        \textcolor{darkgreen}{\# $M$: number of clients}\\
         \textcolor{darkgreen}{\# $\xi$: ($M \times K \times d_p$) learnable prototypes where $\xi[i]$: ($K \times d_p$) corresponds to the set of $K$ prototypes for client $i$}\\
        \textcolor{darkgreen}{\# $\mathcal{N}$: ($M \times M$) adjacency matrix}\\
        \textcolor{darkgreen}{\# $w$: ($M \times M$) weight matrix where $w[i]$: ($1 \times M$) is the locally inferred edge weights of client $i$.}\\
        \textcolor{orange}{def} \hskip0.4em\textcolor{cyan}{\ourmethod{}}():\\
        \hskip1em  \textcolor{orange}{for} round in \textcolor{orange}{range}($T$):\hfill  \textcolor{darkgreen}{\# loop over $T$ global rounds}\\
        \hskip2em  \textcolor{orange}{for} i in \textcolor{orange}{range}($M$):\hfill  \textcolor{darkgreen}{\# loop over all clients}\\
        \hskip3em $\phi_i$, $\xi[i]$ = \textcolor{cyan}{\texttt{PML}}($i$, $\xi[i]$)\hfill  \textcolor{darkgreen}{\# perform a local learning step at client $i$}\\
        \hskip2em \textcolor{orange}{if} $\textup{round} \geq T_{\textup{thr}}$: $w[i]$ = \textcolor{cyan}{\texttt{CGL}}($w[i]$, $\phi_i$, $\Phi = \{\phi_j\}_{j \in \mathcal{N}[i]}$)\hfill \textcolor{darkgreen}{\# update the local weight vector $w_i$ at each round after warmup}\\
        \hskip2em  \textcolor{orange}{for} j in $\mathcal{N}[i]$:\hfill  \textcolor{darkgreen}{\# loop over the set of neighbors}\\
        \hskip3em $\xi[i] = w_i[i]\cdot \xi[i] + w_i[j]\cdot \xi[j]$\hfill  \textcolor{darkgreen}{\# aggregate the set of learnt prototypes over the local neighborhood}\\
\caption{\ourmethod{}: PyTorch-like Psuedocode}
\label{alg:mapl_pytorch}
\end{algorithm}

\begin{algorithm}[ht]
\small
\SetAlgoLined
\DontPrintSemicolon
\SetNoFillComment
        \textcolor{orange}{def} \hskip0.4em\textcolor{cyan}{CGL}($w[i]$, $\phi_i$, $\Phi$):\\
        \hskip1em  \textcolor{orange}{for} j in \textcolor{orange}{range}(\textcolor{orange}{len}($\Phi$)):\hfill  \textcolor{darkgreen}{\# loop over the set of neighbors in $\mathcal{N}_i$}\\    
        \hskip2em s[j] = \textcolor{orange}{cosine\_similarity}($\phi_i$, $\Phi[j]$)\hfill  \textcolor{darkgreen}{\# compute similarity between self and $j^{th}$ client}\\
        \hskip1em s[i] = 1.\hfill \textcolor{darkgreen}{\# assign a score of 1 to self}\\
        \hskip1em s *= -1.\hfill \textcolor{darkgreen}{\# compute the negative of similarity scores}\\
        \hskip1em loss = $\mu_1$ (s * $w[i]$) + $\mu_2$ \textcolor{orange}{regularizer}($w[i]$)\\
        \hskip1em loss.\textcolor{orange}{backwards}()\hfill  \textcolor{darkgreen}{\# compute the gradients}\\
        \hskip1em optimizer.\textcolor{orange}{step}()\hfill  \textcolor{darkgreen}{\# update the weight vector}
\caption{\texttt{CGL}: PyTorch-like Psuedocode}
\label{alg:cgl_pytorch}
\end{algorithm}

\begin{algorithm}[ht]
\small
\SetAlgoLined
\DontPrintSemicolon
\SetNoFillComment
        \textcolor{darkgreen}{\# $\{f_{\theta_i}, f_{\psi_i}, g_{\phi_i}\}$: backbone, projector, and predictor}\\
        \textcolor{darkgreen}{\# E: number of local epochs}\\
        \textcolor{orange}{def} \hskip0.4em\textcolor{cyan}{PML}($i$, $\xi$):\hfill  \textcolor{darkgreen}{\# local step at client $i$}\\
        \hskip1em  \textcolor{orange}{for} iter in \textcolor{orange}{range}(E)\\
        \hskip2em  \textcolor{orange}{for} (x, y) in \textcolor{orange}{loader}($\mathcal{D}_i$)\hfill  \textcolor{darkgreen}{\#$\mathcal{D}_i$: local dataset of client $i$}\\
        \hskip3em  x = [x\_a, x\_b] = [aug1(x), aug2(x)]\hfill  \textcolor{darkgreen}{\# concatenate two augmented views}\\
        \hskip3em z = $f_{\theta_i}$(x)\hfill  \textcolor{darkgreen}{\# ($2B \times d_z$): extract the features}\\
        \hskip3em p = $f_{\psi_i}$(z)\hfill  \textcolor{darkgreen}{\# ($2B \times d_p$): extract the projected features}\\
        \hskip3em o = $g_{\phi_i}$(z)\hfill  \textcolor{darkgreen}{\# ($2B \times k$): obtain the logits.}\\
        \hskip3em loss\_cont = \textcolor{orange}{contrastive\_loss}(p)\hfill  \textcolor{darkgreen}{\# compute the sample-to-sample contrastive loss.}\\
        \hskip3em loss\_proto = \textcolor{orange}{proto\_loss}(p,  $\xi$)\hfill  \textcolor{darkgreen}{\# compute the sample-to-prototype loss.}\\
        \hskip3em loss\_uni = \textcolor{orange}{uniformity\_loss}($\xi$)\hfill  \textcolor{darkgreen}{\# compute the uniformity loss.}\\
        \hskip3em loss\_ce = \textcolor{orange}{cross\_entropy}(o, y.repeat(2))\hfill  \textcolor{darkgreen}{\# compute the cross entropy loss.}\\
        \hskip3em loss = loss\_cont + loss\_proto + loss\_uni + loss\_ce\hfill  \textcolor{darkgreen}{\# total loss computation.}\\
        \hskip3em loss.\textcolor{orange}{backwards}()\hfill  \textcolor{darkgreen}{\# compute the gradients.}\\
        \hskip3em optimizer.\textcolor{orange}{step}()\hfill  \textcolor{darkgreen}{\# update the parameters.}\\
        \hskip1em  \textcolor{orange}{return} $\xi$, $\phi_i$
\caption{\texttt{PML}: PyTorch-like Psuedocode}
\label{alg:pml_pytorch}
\end{algorithm}

\end{document}